\documentclass[acmsmall]{acmart}

\usepackage{algorithm,algorithmic}
\usepackage{graphicx}
\usepackage{textcomp}
\usepackage{xcolor}
\usepackage{bbding}
\usepackage{paralist}
\usepackage{bm}

\usepackage{makecell}
\usepackage{float}
\usepackage{multirow,rotating}
\usepackage{booktabs}
\usepackage{array,threeparttable}
\newcolumntype{P}[1]{>{\centering\arraybackslash}p{#1}}
\usepackage{longtable}

\usepackage{enumitem}

\usepackage{pifont}

\usepackage{forest}

\usepackage{tikz}
\usetikzlibrary{calc,matrix,decorations.markings,decorations.pathreplacing}

\usetikzlibrary{shapes.geometric}
\usetikzlibrary{patterns}
\usetikzlibrary{calc}
\usetikzlibrary{positioning,fit}
\usetikzlibrary{backgrounds}
\usetikzlibrary{intersections}
\usetikzlibrary{shapes,trees,shadows,arrows,chains}

\usepackage{calc}

\AtBeginDocument{%
  }

\setcopyright{acmcopyright}
\copyrightyear{2024}
\acmYear{2024}
\acmDOI{XXXXXXX.XXXXXXX}

\acmJournal{CSUR}
\acmVolume{00}
\acmNumber{0}
\acmArticle{000}
\acmMonth{0}

\begin{document}

\title{Continual Learning on Graphs: A Survey}

\author{ZONGGUI TIAN}
\email{2109853mii30003@student.must.edu.mo}
\author{DU ZHANG}
\authornote{Corresponding}
\email{duzhang@must.edu.mo}
\affiliation{%
  \institution{School of Computer Science and Engineering, Macau University of Science and Technology}
  \streetaddress{Avenida Wai Long, Taipa}
  \city{Macau}
  \country{China}
}

\author{HONG-NING DAI}
\email{henrydai@comp.hkbu.edu.hk}
\affiliation{%
  \institution{Department of Computer Science, Hong Kong Baptist University}
  \streetaddress{55 Renfrew Road, Kowloon Tong}
  \city{HONGKONG}
  \country{China}}

\renewcommand{\shortauthors}{Z.G. Tian et al.}


\begin{abstract}
Recently, continual graph learning has been increasingly adopted for diverse graph-structured data processing tasks in non-stationary environments. Despite its promising learning capability, current studies on continual graph learning mainly focus on mitigating the catastrophic forgetting problem while ignoring continuous performance improvement. To bridge this gap, this article aims to provide a comprehensive survey of recent efforts on continual graph learning. Specifically, we introduce a new taxonomy of continual graph learning from the perspective of overcoming catastrophic forgetting. Moreover, we systematically analyze the challenges of applying these continual graph learning methods in improving performance continuously and then discuss the possible solutions. Finally, we present open issues and future directions pertaining to the development of continual graph learning and discuss how they impact continuous performance improvement.
\end{abstract}




\maketitle 

\section{Introduction}

With the successful applications of deep learning in all walks of life, the community has begun to yearn for more powerful artificial general intelligence. Despite its promising potential, neural network-based continual learning faces a severe forgetting problem: learning on new tasks usually leads to a dramatic performance drop on old tasks, known as catastrophic forgetting (CF)~\cite{prado2022theory}. Continual learning (CL)~\cite{van2019three,hadsell2020embracing} is believed to be a promising way to overcome this challenge. CL is considered a learning ability for intelligent agents to incrementally acquire, update, accumulate, and exploit knowledge to improve their performance on tasks continuously~\cite{hadsell2020embracing}. To alleviate the catastrophic forgetting problem, many CL strategies have been proposed, including replay methods, regularization methods, and parameter isolation methods~\cite{delange2021continual}. These strategies strike a balance between the plasticity and stability of intelligent agents and mitigate the problem of catastrophic forgetting. Nevertheless, current CL only considers individual data samples and ignores the widespread connections between them. Moreover, overcoming CF merely represents an indispensable pathway towards attaining continuous performance improvement (CPI), rather than a terminus of CL.

Graph, also known as network, is a general data representation that describes and analyzes entities with interactions. Graphs have been widely adopted to model diverse types of relations in different applications from biological molecules to social networks. On the one hand, many data naturally exist in the form of graphs, such as citation networks, social networks, and transaction networks. On the other hand, even those data that do not seem to be connected can be artificially constructed into graphs, such as dependency graphs in texts, feature graphs in images, and call graphs in codes. Recently, graph learning has become a promising domain in AI and machine learning due to its strengths in learning intricate relationships among entities and the corresponding network structures. 

However, graph learning has also suffered from the catastrophic forgetting phenomenon. The integration of continual learning with graph learning is apparently also a prevailing solution to mitigate catastrophic forgetting. The consolidation of continual learning with graph learning is called continual graph learning (CGL). Despite CGL's potential, there are significant or intricate differences between general CL and CGL, including models, task settings, and methods due to the structural difference between Euclidean data and graphs. Additionally, both CL and CGL mainly focus on overcoming catastrophic forgetting while ignoring continuous performance improvement. 

Despite an increasing number of research studies on CGL, there are few surveys on CGL. To bridge this gap, this article aims to provide a comprehensive survey of the research efforts in CGL, especially in discussing how the CGL methods achieve continuous performance improvement.

\textbf{Differences between this survey and existing ones.} Since CGL is highly related to continual learning and graph learning, there have been many surveys in both fields. Table~\ref{comparisons} categorizes the related surveys into CL, graph learning, and CGL. In particular, the surveys on continual learning most focus on (i) specific domains, such as natural language processing (NLP)~\cite{biesialska-etal-2020-continual}, computer vision (CV)~\cite{qu2021recent}, robotics~\cite{lesort2020continual}, and autonomous system \cite{shaheen2022continual}); (ii) specific tasks, such as classifications \cite{delange2021continual,mai2022online}); and (iii) models, such as neural networks \cite{parisi2019continual,awasthi2019continual,hadsell2020embracing}). However, they all only consider data from an isolated perspective rather than a consolidated one. Moreover, they over-emphasize the mitigation of catastrophic forgetting while ignoring \emph{continuous performance improvement}, which is the ultimate goal of continual learning.

Regarding the surveys on graph learning, they mainly focus on specific techniques including graph representation learning~\cite{hamilton2017representation,hamilton2020graph,goyal2018graph,cui2018survey,cai2018comprehensive}, graph neural networks~\cite{wu2020comprehensive,zhou2020graph}, and graph deep learning~\cite{zhang2020deep,bacciu2020gentle,georgousis2021graph}. Moreover, most of these studies usually consider data with connections at the sample level while ignoring the connections at the feature level and at the task level. In addition, they only focus on static graphs while ignoring continual learning on dynamic graphs. Although several surveys consider the dynamic nature of graphs, including dynamic graph learning \cite{zhu2022encoder}, dynamic graph representation learning \cite{kazemi2020representation,barros2021survey,xue2022dynamic}, and dynamic graph neural networks \cite{skarding2021foundations}, they mainly consider whether the model adapts to new data while ignoring catastrophic forgetting problem, thereby completely excluding CL. 

To the best of our knowledge, there are only two surveys that comprehensively consolidate continual learning and graph learning. Particularly,~\cite{febrinanto2023graph} reviews the research progress, potential applications, and challenges of CGL while \cite{yuan2023continual} categorizes the methods that overcome catastrophic forgetting in CGL. Although they explicitly consider the connections between data in continual learning and focus on CGL, they do not build a comprehensive perspective and fail to thoroughly elaborate on the relationship and difference between CL and CGL. Furthermore, they mainly focus on mitigating catastrophic forgetting while ignoring continuous performance improvement.

\begin{table}[htbp]
 \begin{threeparttable}
\caption{Comparisons among related existing surveys. }
    \centering
    \begin{tabular}{p{2.5cm}p{3.8cm}p{1cm}<{\centering}p{0.8cm}<{\centering}p{0.8cm}<{\centering}p{0.5cm}<{\centering}p{0.5cm}<{\centering}}
    \toprule
        
    \multirow{2}{*}{References} & \multirow{2}{*}{Focused Scope} & \multicolumn{3}{c}{Graph Perspective} & \multirow{2}{*}{CF} & \multirow{2}{*}{CPI} \\
    
    \cmidrule{3-5}
    
    ~ & ~ & Feature & Sample & Task & ~ \\
    \specialrule{-1em}{1pt}{1pt}
       
    \midrule
        
\cite{wang2023comprehensive} & CL & \ding{55} & \ding{55} & \ding{55} & \ding{51} & \ding{55} \\ 

\cite{cossu2021continual,parisi2019continual,awasthi2019continual,hadsell2020embracing}  & CL in NNs & \ding{55} & \ding{55} & \ding{55} & \ding{51} & \ding{55}  \\ 

\cite{biesialska-etal-2020-continual} & CL in NLP & \ding{55} & \ding{55} & \ding{55} & \ding{51}  & \ding{55} \\

\cite{qu2021recent} & CL in CV & \ding{55} & \ding{55} & \ding{55} & \ding{51} & \ding{55} \\ 
 
\cite{lesort2020continual} & CL in Robotics & \ding{55} & \ding{55} & \ding{55} & \ding{51}  & \ding{55} \\

\cite{delange2021continual,mai2022online} & CL for Classification & \ding{55} & \ding{55} & \ding{55} & \ding{51} & \ding{55}  \\

\cite{shaheen2022continual} & CL in Autonomous System & \ding{55} & \ding{55} & \ding{55} & \ding{51}  & \ding{55} \\
    \hline
    
\cite{xia2021graph,chami2022machine} & GL & \ding{55} & \ding{51} & \ding{55} & \ding{55}  & \ding{55} \\ 

\cite{zhu2022encoder} & DGL & \ding{55} & \ding{51}  & \ding{55} & \ding{55}  & \ding{55} \\

\cite{hamilton2017representation,hamilton2020graph,goyal2018graph,cui2018survey,cai2018comprehensive} & GRL & \ding{55}  & \ding{51}  &  \ding{55} & \ding{55}  & \ding{55} \\

\cite{kazemi2020representation,barros2021survey,xue2022dynamic} & DGRL & \ding{55}  & \ding{51} & \ding{55} & \ding{55}  & \ding{55} \\

\cite{wu2020comprehensive,zhou2020graph} & GNNs & \ding{55}  & \ding{51} & \ding{55} & \ding{55}  & \ding{55} \\

\cite{skarding2021foundations} & DGNNs &  \ding{55} & \ding{51} & \ding{55} & \ding{55}  & \ding{55}  \\

\cite{zhang2020deep,bacciu2020gentle,georgousis2021graph} & GDL & \ding{55} & \ding{51}  & \ding{55} & \ding{55}  & \ding{55} \\
    \hline
                       
\cite{febrinanto2023graph} & CGL & \ding{55}  & \ding{51} & \ding{55} & \ding{51}  & \ding{55} \\ 

\cite{yuan2023continual} & CGL & \ding{55}  & \ding{51} & \ding{51} & \ding{51}  & \ding{55} \\ 

\textbf{This survey} & CGL & \ding{51}  & \ding{51} & \ding{51} & \ding{51}  & \ding{51} \\
    
    \bottomrule
    \end{tabular}
         \begin{tablenotes}
\small
      \item CF: Catastrophic Forgetting; CPI: Continuous Performance Improvement; CL: Continual Learning; (D)GL: (Dynamic) Graph Learning; (D)GRL: (Dynamic) Graph Representation Learning; (D)GNNs: (Dynamic) Graph Neural Networks; GDL: Graph Deep Learning. \ding{51} means considered and \ding{55} means not.
    \end{tablenotes}
\label{comparisons}
   \end{threeparttable}
\end{table}

\begin{figure}[t]
\centering
\includegraphics[scale=0.9]{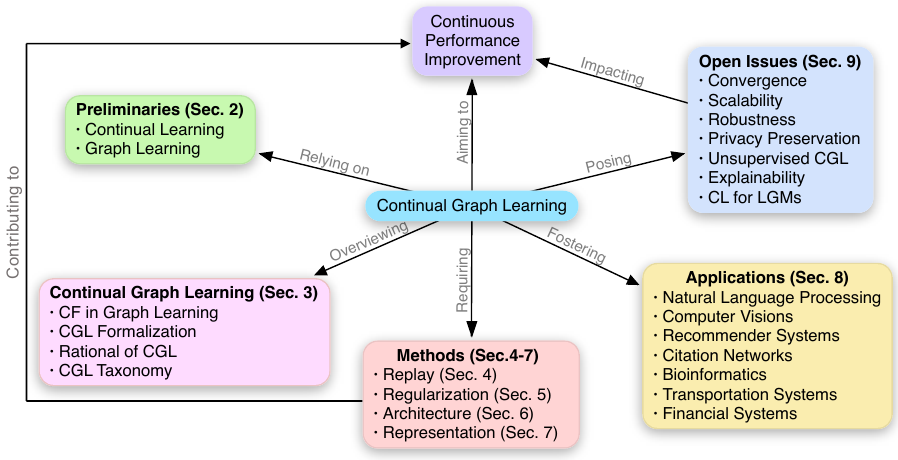}
\caption{Organization of this survey}
\label{organization}
\end{figure}

\textbf{Contributions.} This survey summarizes the state-of-the-art studies on CGL and discusses whether and how current methods can achieve continuous performance improvement. Specifically, our main contributions are summarized as follows: 
\begin{itemize}[wide]
	\item \textbf{A new taxonomy:} We provide a new taxonomy to summarize the methods of overcoming catastrophic forgetting in CGL. Specifically, four groups are introduced from the perspective of how continuous performance improvement can be achieved (see Figure \ref{organization}).
	\item \textbf{A Comprehensive survey:} For each category of method, we discuss the motivation and the main challenges in overcoming catastrophic forgetting. Moreover, we further discuss how the current methods can achieve continuous performance improvement. To the best of our knowledge, this is the first investigation on continuous performance improvement.
	\item \textbf{Future directions:} Focusing on continuous performance improvement, we further propose some related open issues in continual graph learning and discuss how they impact continuous performance improvement, as well as corresponding future directions.
\end{itemize}

Figure~\ref{organization} depicts the organization of this paper. Section~\ref{sec:pre} presents the preliminaries of CL and graph learning. Section~\ref{sec:overview} presents an overview of CGL, including the formalization, motivation, and the new taxonomy of CGL methods that overcome catastrophic forgetting. Specifically, it compares the fields related to CGL from the perspective of specific dimensions. Sections~\ref{sec:replay} to~\ref{sec:rep} summarize the recent advancements of CGL according to the proposed taxonomy. In each category, the main challenges and the corresponding solutions have been surveyed. Moreover, how these methods can achieve continuous performance improvement has also been discussed from the perspectives of knowledge enhancement and optimization controlling, respectively. Section~\ref{sec:app} summarizes real-world applications and the datasets used in existing CLG studies. Thereafter, Section~\ref{sec:open} discusses the open issues and future directions. Finally, Section~\ref{sec:conc} concludes the paper.

\section{Preliminaries}\label{sec:pre}

\subsection{Continual Learning}

Continual learning \cite{van2019three,hadsell2020embracing}, also referred to as lifelong learning \cite{liu2017lifelong,silver2013lifelong}, incremental learning \cite{castro2018end,wu2019large}, never-ending learning \cite{mitchell2018never} or STEP perpetual learning \cite{zhang2018one}, is a machine learning paradigm that continuously learns from time-varying data distributions and learning objectives \cite{lesort2020continual}. In this paradigm, training data usually arrives in batches to complete tasks sequentially. A continual learning system is expected to learn new tasks along the task sequence without forgetting knowledge or experience learned from previous tasks to continuously improve overall performances, such as average accuracy, memory usage, and computation efficiency. An ideal continual learning system is expected to have properties, such as knowledge retention, positive knowledge forward and backward transferability, and online learning capability with task-agnostic and fixed model capacity, according to~\cite{biesialska-etal-2020-continual}. We expand the properties that an ideal continual learning system should have in four dimensions, as summarized in Table~\ref{properties}.

\begin{table}[htbp]
\caption{Ideal continual learning properties}
    \centering
    \begin{tabular}{lp{4cm}p{6.5cm}}
    \toprule
        
    Dimensions & Properties & Descriptions \\ 
        
    \midrule
        
    \multirow{4}{*}{Plasticity
    } & \multirow{2}{*} {Online learning}  & Learning from continuous data stream that may exist data drifts.  \\ 
                       ~ & Task agnostic incremental learning  &  Learning new tasks containing new instances or classes without any prior.  \\ 
    \hline
                       
    \multirow{3}{*}{Stability} & \multirow{2}{*}{Knowledge retention}  & Retain knowledge for continuous performance improvement. \\ 
                       ~ & Learning convergence & The learning process can eventually converge. \\ 
    \hline
    
    \multirow{4}{*}{Transferability} & \multirow{2}{*}{Positive forward transfer} & Previous knowledge can promote follow-up tasks learning. \\ 
        ~ & \multirow{2}{*}{Positive backward transfer} & Follow-up tasks knowledge can improve previous tasks performance. \\
     \hline
                  
    \multirow{4}{*}{Applicability} & \multirow{2}{*}{Causality} & What knowledge led to performance improvement.  \\ 
                      ~ & \multirow{2}{*}{Interpretability} & Why the learned knowledge lead to performance improvement. \\ 
        
    \bottomrule
    \end{tabular}
\label{properties}
\end{table}

Current continual learning systems relax some requirements. For example, most systems are trained in an offline manner, where each batch of data meets the I.I.D assumption, and negative knowledge backward transfer is always inevitable. Large negative backward transfer, known as catastrophic forgetting~\cite{prado2022theory}, is regarded as the main challenge of continual learning.

\subsubsection{\textbf{Continual Learning Scenarios}} 

Currently, the recognized continual learning settings can be categorized into three main scenarios: task incremental continual learning (TICL), domain incremental continual learning (DICL), and class incremental continual learning (CICL). They were proposed following the criteria that the task identity is available or not at test, and adopted in a wide range of studies \cite{van2019three,hsu2018re,zeno2021task}. A more formal definition is introduced in~\cite{hsu2018re} where the three scenarios are formalized through the drift of input and output distributions and whether they share the same representation space.

\begin{itemize}[wide]
	\item TICL: The output spaces are disjoint between tasks and task identities are available in both training and testing.
	\item DICL: Tasks share the same label space while the input data distributions are different. Task identities are neither available in training nor testing.
	\item CICL: The output spaces are disjoint between tasks and task identities are only available in training.
\end{itemize}

More recently, more general and challenging continual learning settings, such as task-free continual learning~\cite{aljundi2019task}, task-agnostic continual learning~\cite{he2020task}, data incremental continual learning~\cite{de2021continual}, data-free continual learning~\cite{smith2021always}, and online continual learning~\cite{lopez2017gradient} have also been proposed, which expands continual learning scenarios.

\subsubsection{\textbf{Continual Learning Methods}}

Continual learning strategies can be mainly classified into three strategies, namely replay-based methods, regularization-based methods, and parameter isolation-based methods, according to \cite{delange2021continual}.

\emph{The replay-based methods} replay part of the historical data saved from previous tasks in the process of learning new tasks to overcome catastrophic forgetting. Many studies based on this strategy have shown good performance \cite{rebuffi2017icarl,rolnick2019experience,isele2018selective,atkinson2021pseudo,shin2017continual,lavda2018continual,ramapuram2020lifelong,lopez2017gradient}. However, since there is an imbalance between the current data and the stored historical mini-batch data, this method usually encounters the problem of biased prediction. Therefore, this strategy is usually used in combination with the regularization-based methods.

\emph{The regularization-based methods} alleviate catastrophic forgetting by imposing constraints on the optimization of the model for new tasks, including adding distillation losses to new models, optimizing and constraining important model parameters, etc. This strategy also showed its applicability in \cite{li2017learning,jung2018less,rannen2017encoder,zhang2020class,kirkpatrick2017overcoming,liu2018rotate,lee2017overcoming,zenke2017continual,aljundi2018memory,chaudhry2018riemannian}. Nevertheless, this strategy suffers from the trade-off between previous knowledge and the performance of new tasks.

\emph{The parameter isolation strategy} uses different parameters for each task to avoid forgetting. Fixed networks mask the networks either at the unit level~\cite{serra2018overcoming} or parameter level \cite{mallya2018packnet,fernando2017pathnet} used for previous tasks when learning a new task. Dynamic architectures add new branches to new tasks and, in the meantime, deactivate previous task parameters~\cite{rusu2016progressive,xu2018reinforced}, or copy models so that independent models are dedicated to each individual task~\cite{aljundi2017expert}. This strategy requires more hyperparameters and more storage.

\subsubsection{\textbf{Evaluation Metrics}}

There are some commonly recognized metrics proposed in~\cite{lopez2017gradient}, including overall performance, backward transfer, and forward transfer.

The \emph{overall performance} of a model reflects its performance on all tasks. It is usually evaluated through average accuracy \cite{lopez2017gradient,diaz2018don} or average incremental accuracy \cite{rebuffi2017icarl,douillard2020podnet,hou2019learning}. Backward transfer and forward transfer reflect the ability of a continual learning system to share knowledge across tasks observed sequentially \cite{prado2022theory}. Specifically, backward transfer is the influence of knowledge learned from new tasks on previous tasks, and forward transfer is the influence of knowledge learned from previous tasks on new tasks \cite{lopez2017gradient}. 

Since both \emph{backward transfer} and \emph{forward transfer} could be either positive or negative, they are usually utilized to evaluate the stability and plasticity, respectively. Additionally, other metrics used to evaluate stability and plasticity are all based on backward transfer and forward transfer. For example, forgetting measure and average forgetting \cite{chaudhry2018riemannian} for stability, and intransigence measure \cite{chaudhry2018riemannian} for plasticity. Current continual learning pursues the trade-off between stability and plasticity.

Suppose that there are $T$ tasks in total and the model has access to the test set of all $T$ tasks. After learning the $i^\text{th}$ task, its test performance on all $T$ tasks is evaluated. Then, a test performance matrix $R \in \mathbb{R}^{T\times T}$ is constructed, and $R_{i,j}$ refers to the test performance of $j^\text{th}$ task after learning on the $i^\text{th}$ task. The above metrics are all evaluated based on the performance matrix $R$. Table~\ref{evaluation} summarizes the main evaluation metrics.

\begin{table}[t]
\caption{Evaluation metrics}
    \centering
    \begin{tabular}{m{2cm}m{3cm}m{6cm}m{1.5cm}}
    \toprule
        
    Metrics & Indicators & Formulas & Source \\ 
        
    \midrule
        
     \multirow{3}{*}{\makecell[l]{Overall \\performance}} & \multirow{2}{*}{Average accuracy} &  \makecell[l]{$ACC_T=\frac{1}{T}\sum_{i=1}^{T}R_{T,i}$\\ $ACC_T^*=\frac{T(T+1)}{2}\sum_{i\geq j}^{T}R_{i,j}$} & \makecell[l]{ \cite{lopez2017gradient} \\ \cite{diaz2018don} } \\ 
     
     ~ &  \makecell[l]{Average incremental \\ accuracy} & \multirow{1}{*}{$AIA_T=\frac{1}{T}\sum_{i=1}^{T}ACC_T$} & \cite{rebuffi2017icarl} \\
                    
    \hline
                       
    \multirow{2}{*}{Stability} & \multirow{2}{*}{Backward transfer}  & \makecell[l]{$BWT=\frac{1}{T-1}\sum_{i=1}^{T-1}R_{T,i}-R_{i,i}$ \\ $BWT^*=\frac{T(T-1)}{2}\sum_{i=2}^T\sum_{j-1}^{i-1}(R_{i,j}-R_{j,j})$} & \makecell[l]{\cite{lopez2017gradient} \\ \cite{diaz2018don} } \\ 
    
    ~ & Average forgetting & $AF_T=\frac{1}{T-1}\sum_{j=1}^{T-1}F_j^T$ & \cite{chaudhry2018riemannian} \\
    
    \hline
    
    \multirow{2}{*}{Plasticity} & Forward transfer &  $FWT_T=\frac{1}{T-1}\sum_{i=2}^{T}(R_{i-1,i}-\widetilde{R}_i)$ & \cite{lopez2017gradient} \\ 
    
    ~ & Intransigence & $IM_T=R^*_T-R_{T,T}$ & \cite{chaudhry2018riemannian} \\
            
    \bottomrule
    \end{tabular}
\label{evaluation}
\end{table}

Besides the above metrics, the evaluation of continual learning models also depends on some specific factors, such as application scenarios and task categories. For example, an online scenario usually has higher requirements on real-time performance, model size, memory efficiency, and robustness. A recommender system may need more metrics than accuracy, like recall or precision. Other metrics, such as model size efficiency, sample storage size efficiency, and computational efficiency were also proposed in~\cite{diaz2018don}.

\subsection{Graph Learning}

Graph learning refers to machine learning on graphs \cite{xia2021graph}. It uses machine learning methods and algorithms to handle graph data to solve analysis tasks on graphs, including classifications, clustering, and predictions.

\subsubsection{\textbf{Graphs and Networks}}

Graphs and networks are usually regarded as equivalent and we use graph and network interchangeably in this paper. A graph is formalized as ${G={  \left\{  V,E \right\}  }}$, where $V={  \left\{  v_1,v_2,...,v_n \right\}  }$ is the node set, and $E={  \left\{ e_{ij}|e_{ij}=(v_i,v_j) \in V \times V \right\}  }$ is the edge set. Generally, graphs can be categorized as static and dynamic, directed and undirected, attributed and unattributed, homogeneous and heterogeneous, homophilic and heterophilic, standard graph, and hypergraph.

\begin{itemize}[wide]
	\item \textbf{Static/Dynamic Graph.} A static graph does not change over time, and a dynamic graph may change on node/edge features and edge connectivities.
\end{itemize}

\begin{itemize}[wide]
	\item \textbf{Directed/Undirected Graph.} A directed graph consists of edges directed from one node to another, i.e., $(v_i,v_j)$ represents the directed edge from vertex $v_i$ to vertex $v_j$. An undirected graph can be considered as a special directed graph where the edges are two-way directed, i.e., $e_{ij}=(v_i,v_j)=(v_j,v_i)=e_{ji}$.
\end{itemize}

\begin{itemize}[wide]
	\item \textbf{Attributed/Unattributed Graph.} An attributed graph refers to a graph where additional information is associated with each vertex or edge, such as labels, and weights. An unattributed graph is a graph in which no additional information is associated with the vertices or edges beyond their connectivity.
\end{itemize}

\begin{itemize}[wide]
	\item \textbf{Homogeneous/Heterogeneous Graph.} A homogeneous graph only has one type of nodes and edges, and a heterogeneous graph consists of multiple types of nodes and edges.
\end{itemize}

\begin{itemize}[wide]
	\item \textbf{Homophilic/Heterophilic Graph.} A homophilic graph is a graph with strong homophily, and a heterophilic graph is a graph with strong heterophily. Homophily/heterophily refers to the property that nodes with similar/different features or same/different class labels are linked together \cite{li2022finding}. Homophily/heterophily is measured by node homophily and edge homophily \cite{zheng2022graph}.
\end{itemize}

\begin{itemize}[wide]
	\item \textbf{Standard Graph/Hypergraph.} A standard graph is a graph where an edge can only connect two nodes, and a hypergraph is a graph where an edge (called hyperedge) can connect any number of nodes \cite{feng2019hypergraph}.
\end{itemize}

\subsubsection{\textbf{Graph Learning Tasks}}

Graph learning tasks can be predictive or generative. Predictive tasks in graph learning involve making predictions or inferences about certain properties or characteristics of the graph, including tasks such as node classification, link prediction, and graph classification. Generative tasks focus on generating new graphs that possess certain desired properties, mainly including graph generation. Overall, for predictive tasks, graph learning tasks can be classified as node-level (e.g. node classification and node regression), edge-level (e.g. link prediction and edge classification), and graph-level (e.g. graph classification and graph regression) based on the level of granularity at which predictions or inferences are made.

\subsubsection{\textbf{Graph Representation Learning}}

Graph representation learning \cite{hamilton2017representation}, also called graph embedding \cite{goyal2018graph} or network embedding \cite{cui2018survey}, aims to learn low-dimensional vector representations for nodes, edges, or entire graphs while capturing the structural and relational information of the graph in these vector representations, feeding downstream graph learning tasks. The widely recognized perspective for graph representation learning is the encoder-decoder framework \cite{hamilton2017representation}, where the learning process is viewed as encoding and decoding.

\begin{itemize}[wide]
	\item \textbf{Graph Encoder.} A graph encoder maps each node in the graph into a low-dimensional vector or embedding, with node features and/or graph structures incorporated optionally.
	\item \textbf{Graph Decoder.} A graph decoder takes the low-dimensional node embeddings and uses them to reconstruct information about each node's neighborhood in the original graph.
\end{itemize}

Note that graph representation learning can be semi-supervised or unsupervised, depending on whether the supervision is available during learning. Specifically, (semi)-supervised methods leverage additional supervision (e.g. node, edge, or graph labels) to learn model parameters. Unsupervised methods learn graph embeddings without labels. 

Graph representation learning can also be performed in transductive and inductive settings according to whether the models can generalize to unseen graph data. In transductive settings, all nodes are observed during training. In inductive settings, models are trained to generalize to new graph data that are not observed during training.

\subsubsection{\textbf{Graph Neural Networks}}

Graph representation learning methods mainly vary with graph encoders, which take different information from the graph and encode it into embeddings. Among these methods, graph neural networks are the mainstream tools. Current GNNs are mainly based on graph convolutions defined in the spatial domain. They mainly follow the idea of message passing to propagate information across nodes in the graph and learn embeddings that encode graph structure and node features. They can also perform graph learning tasks in an end-to-end manner. Additionally, they can be learned in inductive settings. GNNs generally follow the layer propagation formulation as follows,
\begin{gather}
	\mathbf{h}^l_{\mathcal{N}_v} \leftarrow \text{AGGREGATOR}_l\left({\mathbf{h}_u^l,\forall u \in \mathcal{N}_v}\right), \mathbf{h}_{v}^{l+1} \leftarrow \text{UPDATER}_l\left(\mathbf{h}_{v}^l,\mathbf{h}^l_{\mathcal{N}_v}\right),
\end{gather}
\noindent where $\mathbf{h}_u^l$ denotes the representation of node $u$ at the $l^\text{th}$ layer, and $\text{AGGREGATOR}_l$ and $\text{UPDATER}_l$ are the aggregation function and update function at the $l^\text{th}$ layer, respectively. The specific form of the aggregation function varies depending on the GNN architecture, such as sum, attention, mean, or max pooling. The aggregated representations are typically passed through an update function, which can include linear transformations and non-linear activation functions, to obtain the final representation of each node in the $(l+1)^\text{th}$ layer. Representative GNN models include GCN \cite{welling2016semi}, GraphSAGE \cite{hamilton2017inductive}, GAT \cite{velivckovic2017graph}, MPNN \cite{gilmer2017neural}, MoNet \cite{monti2017geometric}, GatedGNN \cite{li2016gated}, and GIN \cite{xu2018powerful}.

\section{Continual Graph Learning}\label{sec:overview}

\subsection{Catastrophic Forgetting in Graph Learning}

Catastrophic forgetting is a common phenomenon in neural networks, and the graph models based on neural networks (mainly GNNs) are no exception. Besides, the imbalance or drift of structure distribution on graphs may also aggravate the catastrophic forgetting problem. \cite{carta2022catastrophic} reports the catastrophic forgetting problem in deep graph networks and confirms the effectiveness of preserving structural information in overcoming catastrophic forgetting.

\subsection{Continual Graph Learning Formalization}

Among the three levels of graph learning tasks, graph-level learning tasks take an entire graph as a sample, and there are no explicit connections between graphs. Therefore, there is no essential difference between graph-level continual graph learning and regular continual learning. For the node-level and edge-level graph learning tasks, each node is considered as a sample, which is interconnected and interdependent. Therefore, the formalization in this section is only applicable for node-level and edge-level continual graph learning tasks.

\begin{figure}[hbtp]
\centering
\includegraphics[scale=0.4]{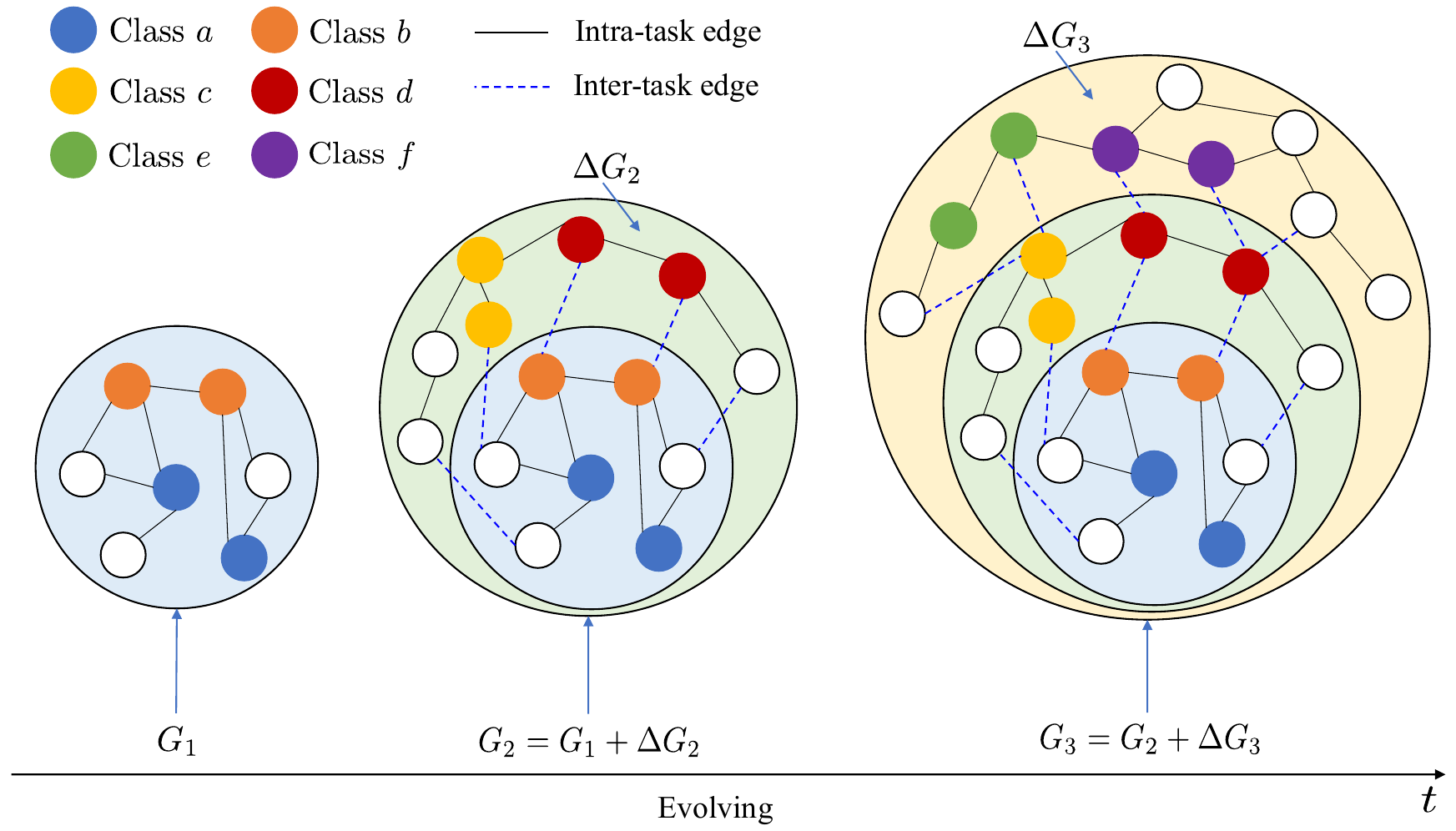}
\caption{An example of a dynamic graph}
\label{dynamic}
\end{figure}

As shown in Figure~\ref{dynamic}, a dynamic graph evolves. In an arbitrary moment $t$, the graph can be observed as a snapshot $G_t$. In this way, a dynamic graph can be represented by a series of snapshots at some specific timestamps.

\textbf{Continual graph learning}: Given a set of snapshots $G_{1:t}={\left\{{G_1,G_2,\cdots,G_t}\right\}}$, where $G_t={\left\{ V_t,E_t \right\}}$ is the snapshot at time $t$. Let $G_1=\Delta G_1$, and $\Delta G_t$ is the change of $G_{t-1}$ in the period from time $(t-1)$ to $t$. Then $G_t=G_{t-1}+\Delta G_t$. Continual graph learning aims to learn a model $f(\Theta)$ from a set of tasks $\mathcal{T}_{1:t}={\left\{ \mathcal{T}_1,\mathcal{T}_2,\cdots,\mathcal{T}_t \right\}}$ corresponding to a set of datasets $\mathcal{D}_{1:t}={\left\{ \Delta G_1,\Delta G_2,\cdots,\Delta G_t \right\}}$ to fit well on all the datasets by minimizing a loss function $\mathcal{L}_\Theta=l(\Theta_t;\mathbf{A}_t,\mathbf{X}_t)$, where $\mathbf{A}_t$ and $\mathbf{X}_t$ are the adjacency matrix of $G_t$ and node feature matrix of $G_t$, respectively, and $\Theta_t$ is the parameters of the graph learning model after training on $G_t$.

\subsection{Rational of Continual Graph Learning}

The most intuitive reason for CGL is the ubiquity of graph-structured data, the time-variability of the world, and the desire for more powerful artificial general intelligence. Specifically, dynamic graph learning and continual learning are closely related to continual graph learning. Table~\ref{comparison} summarizes their connections and differences by considering their properties.

\begin{table}[htbp]
\caption{Comparisons among related areas}
\centering
\resizebox{0.98\linewidth}{!}{
\begin{tabular}{P{2.3cm}P{3.1cm}P{1.6cm}P{1.6cm}P{1.6cm}P{1.6cm}P{1.6cm}}
\toprule
 
 ~ & \multicolumn{6}{c}{Properties} \\ 

\cline{2-7} 

Areas  & Graph representation learning   & Online learning      & Forward transfer     & Backward transfer    & Incremental learning & Knowledge retention  \\

\midrule

Dynamic graph learning   & \multirow{2}{=}{\centering \ding{51}}                              & \multirow{2}{=}{\centering \ding{51}}                     & \multirow{2}{=}{\centering \ding{51}}                     & \multirow{2}{=}{\centering \ding{55}}                  & \multirow{2}{=}{\centering \ding{55}}                     & \multirow{2}{=}{\centering \ding{55}}  \\

\hline

Continual learning       & \multirow{2}{=}{\centering \ding{55}}                               & \multirow{2}{=}{\centering \ding{51}}                     & \multirow{2}{=}{\centering \ding{51}}                    & \multirow{2}{=}{\centering \ding{51}}                   & \multirow{2}{=}{\centering \ding{51}}                     & \multirow{2}{=}{\centering \ding{51}}     \\

\hline

Continual graph learning & \multirow{2}{=}{\centering \ding{51}}                              & \multirow{2}{=}{\centering \ding{51}}                     & \multirow{2}{=}{\centering \ding{51}}                     & \multirow{2}{=}{\centering \ding{51}}                    & \multirow{2}{=}{\centering \ding{51}}                    & \multirow{2}{=}{\centering \ding{51}} \\ 
\bottomrule
\end{tabular}
}
\label{comparison}
\end{table}

\subsubsection*{\textbf{Dynamic Graph Learning VS. Continual Graph Learning}}

Graph learning methods, especially deep learning-based methods like GNNs, have proved their applicability and significance in graph-structured data processing and analysis. \emph{Dynamic graph learning} is proposed to deal with the time-variability of dynamic graphs. It aims to adapt to new snapshots by obtaining the latest graph representations in each time step. In this process, some methods may utilize the learned representations previously to help learn new snapshots, which is referred to as \emph{forward transfer} in continual learning settings. However, it ignores the possible problem of catastrophic forgetting in the process and does not care whether the performance can continue to improve.

\emph{Continual graph learning} differs from dynamic graph learning in that it not only focuses on the forward transfer of knowledge but also on the \emph{backward transfer}. Forward transfer refers to the process of transferring knowledge from one task to another in a task sequence where the learned knowledge from previous tasks is utilized to help subsequent task learning, and backward knowledge transfer is the process of transferring knowledge learned from new tasks to help to improve the performance of previous tasks~\cite{lopez2017gradient}. 

For instance, in a recommender system, users have stable long-term interests but may also be attracted by short-term events. Dynamic graph learning only considers short-term interests and ignores long-term preferences. In contrast, a CGL-based recommender system takes both long and short-term preferences into account jointly to overcome catastrophic forgetting on the former and improve the overall recommender performance continuously. An ideal CGL agent is expected to mitigate catastrophic forgetting and improve performance continuously.

\subsubsection*{\textbf{Continual Learning VS. Continual Graph Learning}}

Current \emph{continual learning} mainly focuses on Euclidian data where the connections are not modeled. A graph describes the ubiquitous connections between data, which are called structures or topologies. Many types of data in the real world can be described as graphs, including social networks~\cite{gao2022graph,fan2019graph}, biological networks \cite{li2021graph,zhang2021graph}, user-item interactions \cite{wu2022graph,tan2020learning}, traffic networks \cite{jiang2022graph,li2021spatial}, and knowledge graphs \cite{zhang2020relational,arora2020survey}. Besides, the features of samples can be also connected in some manners (e.g., class prototypes) to construct a feature graph. Some examples can be referred to HAG-Meta \cite{tan2022graph}, FGN \cite{wang2022lifelong}, RFGP \cite{lei2020class}, and Geometer \cite{lu2022geometer}. Moreover, there are also connections between tasks, which are recognized in multi-task learning and meta-learning. CML-BGNN \cite{luo2020learning} explicitly preserves the  intra- and inter-task correlations. In some cases, a continual graph learning problem can be transformed into a regular continual learning problem (FGN \cite{wang2022lifelong}). 

Therefore, graphs are not the complete opposition of Euclidean data but a supplement of them and provide a connective perspective. CGL not only includes continual learning on graph data but also contains those using graph structural information to perform continual learning on non-graph data. It expands the scope of continual learning by explicitly incorporating the connections between features, samples, and tasks, and preserving them for continual learning. These connections are mainly modeled through graph representation learning. On the other hand, it also brings new challenges and opportunities (new methodology to overcome catastrophic forgetting) in the preservation and fusion of these connections. Whether these connections can enhance the model performance or help for continual performance improvement still needs more explorations and evidence.

\subsection{Continual Graph Learning Taxonomy}

Continual graph learning is essentially a sub-field of continual learning, and therefore, the aim of continual graph learning is the same as regular continual learning: continuous performance improvement of models through incremental learning. Nevertheless, due to the interdependence between nodes in graphs, continual graph learning is more complicated and diverse in methods than that of regular one. Therefore, we propose a new taxonomy of current continual graph learning methods, which falls into four categories: replay-based methods, regularization-based methods, architecture-based methods, and representation-based methods, as shown in Figure~\ref{taxonomy}.

\tikzset{
  basic/.style   = {draw, text width=5cm, drop shadow, font=\sffamily, rectangle},
  root/.style    = {basic, rounded corners=2pt, thin, align=center, fill=white},
  level-2/.style = {basic, rounded corners=6pt, thin, align=center, fill=white, text width=3cm},
  level-3/.style = {basic, thin, align=flush center, fill=white, text width=2.4cm}
}

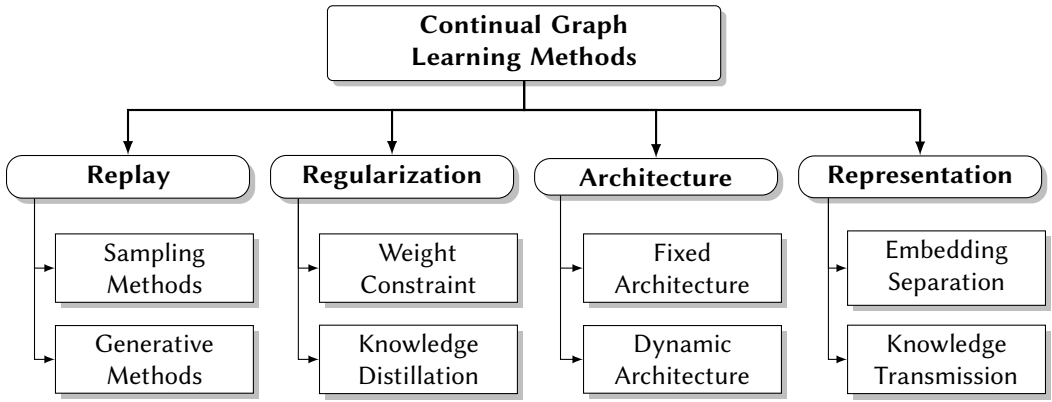
\begin{figure}[tbp]
    \centering
\begin{tikzpicture}[
  level 1/.style={sibling distance=10em, level distance=5em},
  edge from parent/.style={->,solid,black,thick,sloped,draw}, 
  edge from parent path={(\tikzparentnode.south) -- (\tikzchildnode.north)},
  >=latex, node distance=1.2cm, edge from parent fork down]

\node[root] {\textbf{Continual Graph Learning Methods}}
  child {node[level-2] (c1) {\textbf{Replay}}}
  child {node[level-2] (c2) {\textbf{Regularization}}}
  child {node[level-2] (c3) {\textbf{Architecture}}}
  child {node[level-2] (c4) {\textbf{Representation}}};

\begin{scope}[every node/.style={level-3}]
\node [below of = c1, xshift=10pt] (c11) {Sampling Methods};
\node [below of = c11] (c12) {Generative Methods};

\node [below of = c2, xshift=10pt] (c21) {Weight Constraint};
\node [below of = c21] (c22) {Knowledge Distillation};

\node [below of = c3, xshift=10pt] (c31) {Fixed Architecture};
\node [below of = c31] (c32) {Dynamic Architecture};

\node [below of = c4, xshift=10pt] (c41) {Embedding Separation};
\node [below of = c41] (c42) {Knowledge Transmission};
\end{scope}

\foreach \value in {1,2}
  \draw[->] (c1.193) |- (c1\value.west);

\foreach \value in {1,2}
  \draw[->] (c2.193) |- (c2\value.west);

\foreach \value in {1,2}
  \draw[->] (c3.191) |- (c3\value.west);
  
\foreach \value in {1,2}
  \draw[->] (c4.193) |- (c4\value.west);
\end{tikzpicture}
    \caption{A taxonomy of continual graph learning methods}
    \label{taxonomy}
    \vspace{-0.8cm}
\end{figure}

The proposed taxonomy mainly focuses on the methods that explicitly claim to overcome catastrophic forgetting since the current work rarely involves continuous performance improvement. Nevertheless, we discuss the root cause of catastrophic forgetting and the key to continuous performance improvement from the perspective of knowledge and further discuss whether and how these methods can achieve continuous performance improvement.

From the perspective of knowledge, the root cause of catastrophic forgetting is the coverage of new knowledge on existing knowledge. Assuming that the knowledge of a certain period of time is finite and it can be learned in continual learning settings, then the goal of continual learning is to learn all the knowledge and achieve continuous performance improvement on specific tasks. In light of this consideration, continuous performance improvement is equivalent to the continuous acquisition of new knowledge or the complement of existing knowledge. It can generally achieved in two ways: knowledge enhancement and optimization control.

Knowledge enhancement refers to the knowledge of subsequent tasks that can enhance the knowledge of previous tasks. For example, humans learn four basic math operations in lower grades and use them to solve real-world questions. However, they do not use variables to represent numbers and thus easily make mistakes in understanding the quantitative relationships between objects. After they learn variables and equations in their senior years, they will use variable equations to understand and model the quantitative relationships between objects, which gives them less chance to make mistakes. In this example, the variables and equations are the enhancement to the basic four math operations. Knowledge enhancement can be either achieved through learning positive samples or negative samples. 

Optimization controlling refers to controlling the learning process. If the learning process can be quantified with the degree of completeness, complete learning definitely outperforms incomplete learning. In analogy, students who listen carefully in classes and complete all homework usually outperform their counterparts who distract from classes and leave their homework blank. 

In this paper, we follow the above considerations to discuss and analyze whether and how the current continual graph learning methods can achieve continuous performance improvement.

\section{Replay-based Methods}\label{sec:replay}

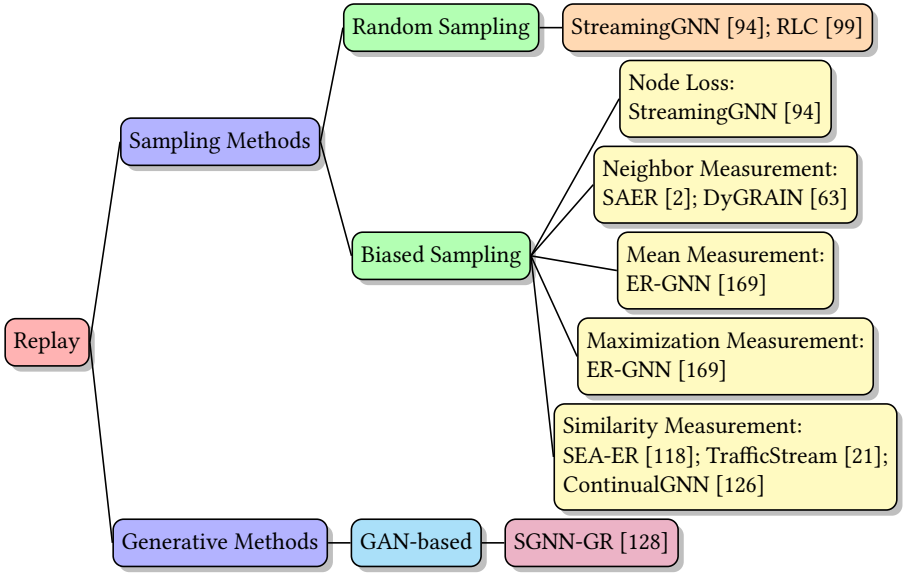
\begin{figure}[tbp]
\centering
\begin{forest}
    for tree={
      draw, semithick, rounded corners, drop shadow,
            align = left,   
             edge = {draw, semithick},
    parent anchor = east, 
     child anchor = west,
     		 grow = 0, reversed, 
            s sep = 1mm,    
            l sep = 3mm,    
            font  = \small
               }
[Replay, fill=red!30
    [Sampling Methods, fill=blue!30
	[Random Sampling, fill=green!30
		[StreamingGNN \cite{perini2022learning}; RLC \cite{rakaraddi2022reinforced}, fill=orange!30]
	]
        [Biased Sampling, fill=green!30
            [Node Loss: \\ StreamingGNN \cite{perini2022learning}, fill=yellow!30]
            [Neighbor Measurement: \\ SAER \cite{ahrabian2021structure}; DyGRAIN \cite{kim2022dygrain}, fill=yellow!30]
            [Mean Measurement: \\ ER-GNN \cite{zhou2021overcoming}, fill=yellow!30]
            [Maximization Measurement: \\ ER-GNN \cite{zhou2021overcoming}, fill=yellow!30]
            [Similarity Measurement: \\ SEA-ER \cite{su2023towards}; TrafficStream \cite{chen2021trafficstream}; \\ ContinualGNN \cite{wang2020streaming}, fill=yellow!30]
        ]
    ]
    [Generative Methods, fill=blue!30
        [GAN-based, fill=cyan!30
            [SGNN-GR \cite{wang2022streaming}, fill=purple!30]
        ]
    ]
]
\end{forest}
\caption{Replay-based CGL methods}
\label{replay}
\end{figure}

The replay-based methods leverage the knowledge learned from previous tasks for joint training with current data when learning new tasks to avoid catastrophic forgetting. The key to replay-based methods is the acquisition of the knowledge learned from previous tasks, which is generally obtained by sampling or generative models. Figure~\ref{replay} summarizes the replay-based methods.

\subsection{Sampling Methods}

Sampling methods select some historical data from previous tasks and store them in the buffer to preserve the learned knowledge for overcoming catastrophic forgetting. Due to the limited buffer space, how an appropriate sampling strategy is developed to make effective use of the buffer is a challenge. Specifically, it involves the construction and updating of the buffer. For buffer construction, a sampling strategy should specify the buffer size and the range of the sampling carried out. For the buffer updating, the sampling strategy needs to specify what is the criteria for the node selection and how the criteria are evaluated quantitatively.
 
\subsubsection{\textbf{Random Sampling}} 

The most simple strategy is random sampling where each node is selected with equal probabilities. StreamingGNN \cite{perini2022learning} mentions a uniform random sampling that selects a fixed-size sample of the graph seen so far for training with the new data. However, this sampling can only work well for the i.i.d. data and cannot handle the unbalanced data. Another strategy to handle complex structural dependencies is biased sampling where nodes are selected with different probabilities. The higher the importance of the node, the greater the probability of being selected. Therefore, the crucial issue is the measurement of the node's importance. RLC \cite{rakaraddi2022reinforced} follows the reservoir sampling strategy to select nodes randomly with equal probabilities.

\subsubsection{\textbf{Biased Sampling}}

Biased sampling considers nodes with different importance levels, which can be defined with several measurements, including node loss, simple neighborhood, mean, maximization, and similarity.

StreamingGNN \cite{perini2022learning} proposes to compute the loss of nodes to determine their priorities. The probability of being sampled is proportional to the prediction errors of nodes. This strategy is based on the idea that the errors of rare nodes decrease slower than those of common ones, and more selections on rare nodes lead to less forgetting.

Neighbor measurement follows an intuitive and straightforward idea that the importance of a node is determined by its neighborhood. SAER \cite{ahrabian2021structure} believes that the nodes with a smaller number of interactions are less informative than that of large ones. Therefore, it proposes that the probability of each interaction is proportional to the inverse degree of its user node to prevent information loss. DyGRAIN \cite{kim2022dygrain} notes the time-varying receptive field and defines the influence propagation in view of structure and feature. It introduces propagation and influence matrix and corresponding score to measure the influence of new data on each node. A feature influence score is proposed based on the node's feature vector to capture the amount of change in the receptive field of nodes. The nodes with high scores are selected as important for replay. 

Mean and maximization measurements are proposed in ER-GNN \cite{zhou2021overcoming}. A simple way to identify important nodes is to adopt a fixed size of nodes whose embeddings are closest to the mean of the feature of each class, called the mean of the feature. However, this approach is limited by the number of nodes in each class. Coverage maximization is proposed for the limitation of a small number of nodes and it aims to maximize the coverage of embedding space, which is proportional to the number of nodes from other classes in the same task within a fixed distance. Influence maximization takes the parameter changes as the metric to measure the importance of nodes, where the parameter changes are evaluated by comparing the parameters obtained by removing a node and the optimal parameters obtained by not removing it. The greater the change, the more important the node is. Since the influence calculation of every removed training node is highly expensive, it converts the removal of a training node to upweight it on the optimal parameters. 

Similarity measurement is a common way to determine node importance and it can be defined in several ways, including structural similarity, attributes or features similarity, and distribution similarity. SEA-ER \cite{su2023towards} considers that the replay nodes should be able to generalize well to other nodes. Therefore, it proposes to select the nodes that have the closest structural similarity to other nodes. Specifically, the structural similarity is evaluated by the short path distance from a node to other nodes. ContinualGNN \cite{wang2020streaming} regards a node as important if its attributes are significantly different from that of its neighbors. The more the node number of different labels in its neighbors, the more important the node is. TrafficStream \cite{chen2021trafficstream} proposes to detect greatly changed nodes based on the distribution discrepancy of the output features in different time intervals. The distribution discrepancy is measured by the Jensen-Shannon divergence and the greater value means more substantial changes and more influenced nodes. The lower scores reveal that the data flow is more stable and can be chosen as a replay.

\subsection{Generative Methods}

Generative methods generate pseudo-data of previous tasks through generative models to preserve old knowledge for incorporation with new data. The key to this method is the generative model itself. Specifically, there are two issues: how can the pseudo-data generated by the generative model effectively represent the previous knowledge? How does the generative model update with the incremental tasks?

SGNN-GR \cite{wang2022streaming} proposes a generative model based on the GAN framework to generate node neighborhoods. It converts the learning problem of the node neighborhood into the learning of random walks with a restart on graphs. Compared with a simple random walk, it permits to full exploration of the node neighborhood within a small number of steps, which ensures the good performance of the generative model.

\subsection{Summary}

Corresponding to the discussion above, we briefly summarize the replay-based continual graph learning methods and discuss how they can realize continuous performance improvement.

\subsubsection{\textbf{Sampling}} Typically, a buffer for the sampling methods is with a fixed size. Random sampling is the most simple way where each node seen so far is sampled with equal probability. However, this sampling strategy ignores node importance and therefore cannot update the buffer efficiently. Therefore, as the tasks increase, random sampling either samples historical nodes through multiple repetitions to construct the buffer or randomly replaces the nodes in the buffer to update it. 

Biased sampling considers the differences between nodes and gives them importance based on certain rules, such as feature mean, node loss, node degree, influence, coverage, structural similarity, and attribute similarity. They also follow the fixed-size buffer principle and are usually developed based on Reservoir sampling. They usually outperform random sampling due to more involved strategies. For different biased sampling methods, an important question is whether they can be sensitive to changes in the topological structures and update the node importance according to such changes efficiently. Another question is whether the sampling based on the importance of updating strategy can effectively retain the knowledge of preceding tasks, so as to effectively overcome catastrophic forgetting and even achieve continuous performance improvement.

\subsubsection{\textbf{Generative}} Generative methods involve the generative model itself. Under the settings of continual graph learning, an ideal generative model should be able to update incrementally and generate better pseudo-data as the tasks increase to preserve previous knowledge better. Many deep graph generative models have been proposed so far. They are roughly classified into GAN-based (e.g., MolGAN \cite{de2018molgan}), VAE-based (e.g., GraphVAE \cite{simonovsky2018graphvae}), flow-based (e.g., GraphAF \cite{shi2019graphaf}), reinforcement learning-based (e.g., GCPN \cite{you2018graph}), and autoregressive-based (e.g., GraphRNN \cite{you2018graphrnn} and BiGG \cite{dai2020scalable}). Current work SGNN-GR \cite{wang2022streaming} only focuses on GAN-based models, leaving a broad space for exploration of this method.

\subsubsection{\textbf{Continuous Performance Improvement}}

Since replay-based methods only store partial historical data, which can transfer a portion of the existing knowledge to the next tasks, but not the new knowledge of the next tasks to previous tasks. There is no explicit knowledge enhancement in this process and thus cannot achieve continuous performance improvement theoretically. 

Nevertheless, the model performance is jointly determined by many factors such as data quality, data volume, task difficulty, and algorithm ability. After clarifying tasks and selecting algorithms, data quality and volume are the determinants of performance. If the knowledge contained in the subsequent tasks like data distributions is consistent with or similar to that of the previous tasks, there may also be opportunities for continuous performance improvement. Other ways to this possibility may also include some new scenarios combined with other learning paradigms, like multi-modality learning, multi-view learning, multi-label learning, and multi-supervision learning, where the data of previous tasks are enhanced with new modalities, views, labels, and supervisions (from weak to strong, supervised to unsupervised, etc.).

Besides, continuous performance improvement may be achieved through optimization controlling by imposing constraints, as proposed in GEM \cite{lopez2017gradient}, which constrains the losses of previous tasks not increasing. Other ways may include various regularization and optimization methods like early stopping, dropout, etc.

\section{Regularization-based Methods}\label{sec:reg}

\begin{figure}[t]
\centering
\begin{forest}
    for tree={
      draw, semithick, rounded corners, drop shadow,
            align = left,   
             edge = {draw, semithick},
    parent anchor = east, 
     child anchor = west,
     		 grow = 0, reversed, 
            s sep = 1mm,    
            l sep = 3mm,    
            font=\small
               }
[Regularization, fill=red!30
    [Weight Constraint, fill=blue!30
	[Topology Representation, fill=green!30
		[Local structure: \\ TWP \cite{liu2021overcoming}, fill=orange!30]
		[Node representation: \\TrafficStream \cite{chen2021trafficstream}; \\ ContinualGNN \cite{wang2020streaming}, fill=orange!30]
		[Model architecture: \\ MSCGL \cite{cai2022multimodal}, fill=orange!30]
	]
        [Importance Measurement, fill=green!30
            [Gradient: TWP \cite{liu2021overcoming}, fill=yellow!30]
            [Fisher information matrix: \\ TrafficStream \cite{chen2021trafficstream}; \\ ContinualGNN \cite{wang2020streaming}, fill=yellow!30]
            [Orthogonal space: MSCGL \cite{cai2022multimodal}, fill=yellow!30]
        ]
    ]   	
    [Knowledge Distillation, fill=blue!30
        [Knowledge Type, fill=cyan!30
            [Logits: \\ Geometer \cite{lu2022geometer}, fill=purple!30]
            [Topology: \\ ERDIL \cite{dong2021few}; SKGD \cite{liu2021structural}; \\ GraphSAIL \cite{xu2020graphsail}, fill=purple!30]
            [Embeddings: \\ DyGRAIN \cite{kim2022dygrain}; SKGD \cite{liu2021structural}; \\ LWC-KD \cite{wang2021graph}; GraphSAIL \cite{xu2020graphsail}; \\ RieGrace \cite{sun2023self}, fill=purple!30]
        ]
        [Distillation Paradigm, fill=cyan!30
            [Vanilla Distillation: \\ DyGRAIN \cite{kim2022dygrain}; ERDIL \cite{dong2021few}; \\ Geometer \cite{lu2022geometer}; GraphSAIL \cite{xu2020graphsail}; \\ SKGD \cite{liu2021structural}, fill=pink!30]
            [Contrastive Distillation: \\ LWC-KD \cite{wang2021graph}; RieGrace \cite{sun2023self}, fill=pink!30]
        ]
    ]
]
\end{forest}
\caption{Regularization-based CGL methods}
\label{regularization}
\end{figure}
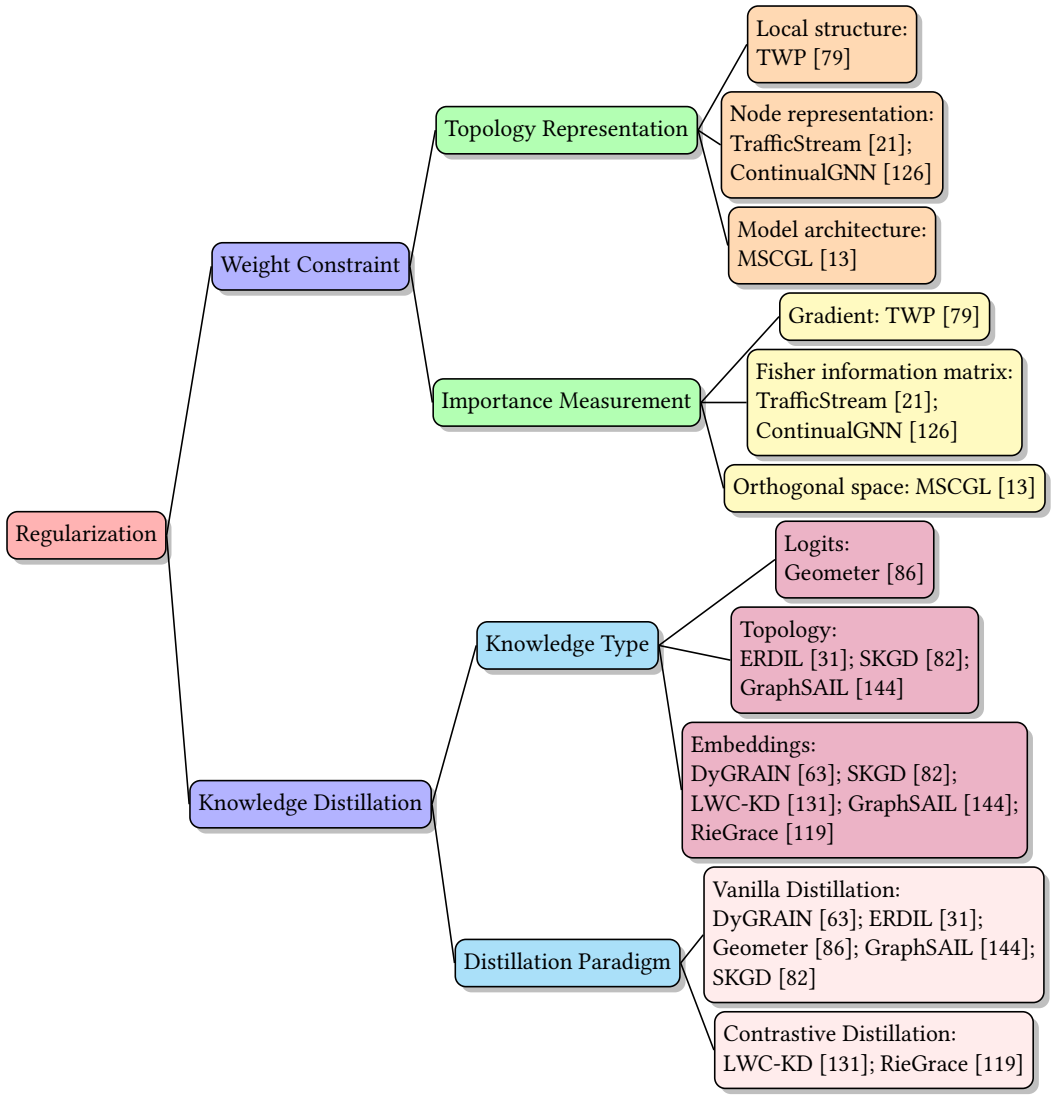

The regularization-based methods balance the learning of old and new tasks by explicitly considering the topological structures and adding corresponding regularization terms to the loss function to regularize the gradient directions, and therefore limit the drastic changes of parameters that are significant to previous tasks to overcome catastrophic forgetting. The regularization terms usually come in two ways: constrain and distillation. Figure~\ref{regularization} summarizes the regularization-based methods.

\subsection{Weight Constraint}

Weight constraint methods limit the learning of new tasks and retain old knowledge by penalizing the update of parameters that are important to old tasks. Since the topological structures are considered, there are two key issues: what knowledge can represent the topological structures and how they are encoded, and which parameters are important and how the importance is measured? They are referred to as topology representation and importance measurement.

\subsubsection{\textbf{Topology Representation}}

It is essential to consider topological information in continual graph learning tasks. In most cases, considering the local structure of a graph is capable of training effective GNNs because they usually perform best when using two or three layers of graph convolution due to the over-smoothing problem. A general way to encode the topological information is the attention between nodes. TWP \cite{liu2021overcoming} is a typical practice of the principle of attention-as-the-correlations. It also generalizes the attention coefficients to general GNNs by introducing a non-parametric attention mechanism.

The above two works explicitly consider the local structures as the topology information and integrate them into the deep graph models. However, in many cases, the topology information is not explicitly encoded. ContinualGNN \cite{wang2020streaming} and TrafficStream \cite{chen2021trafficstream} both identify the 2-hops neighbors of the newly added nodes as the influenced nodes. The structural information is considered to be included in the changes in the representations of the influenced nodes.

Besides the changes of node representations, the model architectures also serve as the carrier of knowledge, just as the practice in MSCGL \cite{cai2022multimodal} where the obsolete blocks of old architectures are removed and new blocks are added according to the results obtained by the neural architecture search.

\subsubsection{\textbf{Importance Measurement}}

There are several ways to measure the parameter importance. A common way is through the Fisher information matrix of the previous task to estimate the parameter importance, just as the practice in ContinualGNN \cite{wang2020streaming} and TrafficStream \cite{chen2021trafficstream} both use the Fisher information matrix. A diagonal approximation of the Fisher information matrix is usually adopted for computational efficiency.

Since parameter updating is determined by gradient direction, TWP \cite{liu2021overcoming} defines the gradient of the attention coefficient with respect to a specific parameter as its importance. It also considers the task-specific objective as important knowledge and defines the gradient of the task-specific loss with respect to a specific parameter as its importance. The final score is the weighted sum of both importance levels.



A simple way to find important parameters is that all the parameters of the previous task are considered important and the parameters for learning new tasks are updated following the orthogonal direction of the previous parameter space. MSCGL~\cite{cai2022multimodal} search best task-specific architectures and parameters. The parameters that are orthogonal to that of the previous task are encouraged to be updated. 

\subsection{Knowledge Distillation}

Knowledge distillation in continual graph learning problems mainly aims at performance improvement rather than model compression. Constraint methods regularize the models in parameter space, while distillation methods regularize the functional space of the models. Therefore, distillation methods can prevent the model from deviating from what it previously learned to preserve old knowledge. The previously learned model usually serves as the teacher and the currently learned model is as the student. It can be regarded as the self-distillation paradigm compared with the traditional two-step distillation schemes. There are two key issues for distillation methods. What knowledge does the teacher want the student to learn (what to distill)? How is knowledge transferred from the teacher to the student through distillation (how to distill)? They are referred to as knowledge type and distillation paradigm. 

\subsubsection{\textbf{Knowledge Type}}

In early knowledge distillation, logits-based knowledge is widely used for distillation, which also exists in graph deep models. Geometer \cite{lu2022geometer} calculates the KL-divergence of the softened logits to enable the student model to learn from the teacher model to classify old classes.

The topological information in graphs is also a significant knowledge source, and some work \cite{dong2021few,xu2020graphsail,liu2021structural} explicitly preserve topological information for continual learning. ERDIL \cite{dong2021few} considers the angle relations in exemplar relation graphs as important knowledge and leverages the distillation to transfer the structural information for continual learning. GraphSAIL \cite{xu2020graphsail} considers the local and global structural information significant in recommender systems and preserves them through knowledge distillation. SKGD~\cite{liu2021structural} explicitly retains the edge weights as the structural information of the memory knowledge graph for distillation. 

Besides the logits-based and topology-based knowledge, feature-based knowledge is another knowledge used for distillation where the student models learn middle-layer features from teacher models. Representative work includes DyGRAIN \cite{kim2022dygrain}, LWC-KD \cite{wang2021graph}, GraphSAIL \cite{xu2020graphsail}, RieGrace \cite{sun2023self}, and SKGD \cite{liu2021structural}. They all follow the teacher-student architecture and use the information based on embeddings as knowledge for distillation, including middle-layer embeddings, the mutual information between the teacher and student representations, the center node and its neighbor embeddings, low-level and high-level features, and semantic features.

\subsubsection{\textbf{Distillation Paradigm}}

The distillation paradigm specifies how knowledge is distilled, which is closely related to the distillation architecture and distillation objectives. The basic distillation paradigm is vanilla distillation, which follows the general teacher-student architecture and uses the predictions of the teacher model as the supervision to instruct the learning of the student model. Most work follows the vanilla distillation, including DyGRAIN \cite{kim2022dygrain}, ERDIL \cite{dong2021few}, Geometer \cite{lu2022geometer}, GraphSAIL \cite{xu2020graphsail}, and SKGD \cite{liu2021structural}.
 
On the other side, in contrastive distillation, the student model does not directly imitate the teacher model's predictions but is trained to distinguish between the correct class and other incorrect classes, which is performed by generating contrastive examples. The student model learns to differentiate between pairs of similar inputs with different labels, which enables the model to capture more fine-grained details and improve its generalization capabilities. LWC-KD \cite{wang2021graph}, and RieGrace \cite{sun2023self} are the typical work using contrastive distillation.

\subsection{Summary}

This part briefly summarizes the reviewed works in terms of two ways of overcoming catastrophic forgetting and the possible ways to continuous performance improvement.

\subsubsection{\textbf{Weight Constraint}}

Essentially, weight constraint methods in continual graph learning are the same as those in general continual learning, which both limit the updating of important parameters to previous tasks. The difference is that in continual graph learning, the topological information is also significant to task learning. Current work incorporates topological information into the continual learning process explicitly or implicitly through various ways, like attentions \cite{liu2021overcoming}, node representations \cite{wang2020streaming,chen2021trafficstream}, and also architecture \cite{cai2022multimodal}. 

Correspondingly, the parameter importance measurement not only considers the importance of learning tasks but also takes the importance of preserving topological information into account. The measurement includes gradient, Fisher information matrix, and orthogonal projection. However, most current work implicitly incorporates topological information, which brings efficiency issues for topology representation and parameter importance estimation.

\subsubsection{\textbf{Knowledge Distillation}}

Knowledge distillation methods mainly focus on knowledge type and distillation paradigm. Besides the general logits and embedding-based knowledge, topology-based knowledge is a significant consideration for distillation in continual graph learning, like local structure, global structure, relation structure, etc. A key issue is to clarify what kind of topology-based knowledge is important for task learning.

Most studies follow the vanilla distillation paradigm and a few of them perform contrastive distillation. Contrastive learning learns more robust representations by maximizing feature consistency under differently augmented views by exploiting data- or task-specific augmentations \cite{you2020graph}, which enables continual learning more likely to achieve less forgetting and even continuous performance improvement.

\subsubsection{\textbf{Continuous Performance Improvement}}

Regularization-based methods usually formulate continual learning as a posterior distribution approximation problem under the Bayesian framework where the previous data distribution is regarded as the prior knowledge and is propagated to the next task. Essentially, regularization-based methods are established on the basis of model over-parameterization where most parameters rarely contribute to the tasks and therefore can be set as approximate zero or pruned. However, due to the inevitable updating interference between parameters, it is usually unavoidable to strike a trade-off in performances between the previous tasks and new tasks. Some work \cite{chaudhry2020continual,farajtabar2020orthogonal,guo2022adaptive} use gradient orthogonal projection to selectively update parameters to minimize the impact of new task learning on the parameters of previous tasks, which can reduce forgetting to a very low level. Nevertheless, similar to replay-based methods, there is no knowledge enhancement in this process and it can hardly achieve continuous performance improvement theoretically. 
 
A possible way is knowledge distillation. Some work \cite{xie2020self,wang2021knowledge,romero2015fitnets} show that student models can outperform teacher models in certain scenarios, especially when the teacher model is over-parameterized or too complex for the task at hand. They are mostly achieved by mutual learning, self-training, and ensemble learning, which enhance knowledge in various ways, and therefore, it is possible to achieve continuous performance improvement in continual graph learning settings.

\section{Architecture-based Methods}\label{sec:arch}

\begin{figure}[tbp]
\centering
\begin{forest}
    for tree={
      draw, semithick, rounded corners, drop shadow,
            align = left,   
             edge = {draw, semithick},
    parent anchor = east, 
     child anchor = west,
     		 grow = 0, reversed, 
            s sep = 1mm,    
            l sep = 3mm,    
            font  = \small
               }
[Architecture, fill=red!30
    [Fixed Architecture, fill=blue!30
	[Parameter Allocation, fill=green!30
		[Neural Architecture Search: \\ MSCGL \cite{cai2022multimodal}, fill=orange!30]
	]
        [Parameter Efficiency, fill=green!30
            [Parameter Sharing: \\ MSCGL \cite{cai2022multimodal}, fill=yellow!30]
            [Group Sparse Regularization: \\ MSCGL \cite{cai2022multimodal}, fill=yellow!30]
        ]
    ]   	
    [Dynamic Architecture, fill=blue!30
	[Architecture Learning, fill=cyan!30
		[Reinforcement Learning: \\ RLC \cite{rakaraddi2022reinforced}, fill=purple!30]
	]
        [Parameter Efficiency, fill=cyan!30
            [Pruning: \\ RLC \cite{rakaraddi2022reinforced}, fill=pink!30]
        ]
    ]
]
\end{forest}
\caption{Architecture-based CGL methods}
\label{architecture}
\end{figure}
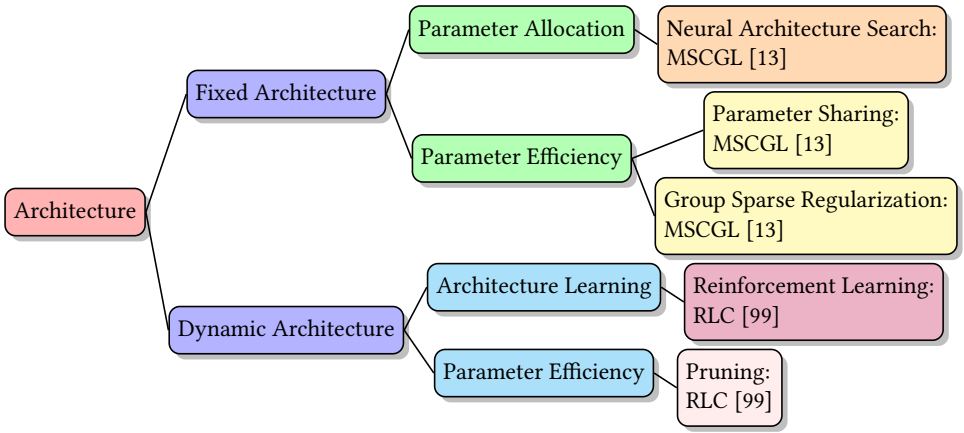

The architecture-based methods assign task-specific parameters or networks (partially or not shared) for tasks through specific architectures to avoid interference between tasks. The architectures can be fixed or dynamic, as shown in Figure \ref{architecture}.

\subsection{Fixed Architecture}

Neural networks with fixed architecture assign specific parameters or neurons to each task and freeze them that belong to old tasks when learning new tasks to avoid catastrophic forgetting. Since fixed architecture limits the model capacity, the parameters or neurons need to be partially shared between tasks as the new tasks are introduced incrementally. There are two main issues for parameter reuse: parameter allocation or architecture learning (how the optimal parameters/architectures are obtained), and parameter efficiency (how the efficiency of parameter reuse is ensured, i.e., maximum sharing of architecture and parameters).

\subsubsection{\textbf{Parameter Allocation}}

Parameter allocation specifies how the model parameters are assigned to tasks. MSCGL \cite{cai2022multimodal} adopts a neural architecture search that employs reinforcement learning to search the optimal network architectures. It formulates the process of searching as an optimization problem and trains a controller to search for model architecture and weights jointly.

\subsubsection{\textbf{Parameter Efficiency}}

To ensure model scalability, the fixed architecture methods need to consider the parameter efficiency, i.e., how the parameters are reused or shared between tasks. A typical solution is a regularization that imposes sparse constraints on parameters. MSCGL \cite{cai2022multimodal} imposes group sparse regularization on the parameter space to search for the parameters that are both block sparse and orthogonal to previously shared parameters. Besides, it also explores several parameter sharing strategies, which share parameters between the same GNN cells, same aggregations, and all aggregations, which also improves parameter efficiency.

\subsection{Dynamic Architecture}

When the capacity of the model with a fixed architecture reaches the learning limit, the dynamic architecture model is an alternative to the fixed one for continual learning. Since the architecture is expanding as the tasks are introduced incrementally, it is generally required to search for the best architecture, which is referred as architecture learning. Besides, the sparsity constraints are also required for scalability.

\subsubsection{\textbf{Architecture Learning}}

Architecture learning is usually achieved by reinforcement learning, as the practice in RLC \cite{rakaraddi2022reinforced}. It consists of a reinforcement learning-based controller and a base trainable network child network. The reinforcement learning-based controller which consists of the LTSM network aims to learn the optimal state policy to select the appropriate course of actions from the user-defined search space on each trainable layer of the CN.

\subsubsection{\textbf{Parameter Efficiency}}

Dynamic architecture also needs to consider parameter efficiency due to the model expansion should be much slower than the task increase to ensure scalability. RLC \cite{rakaraddi2022reinforced} adopts pruning to delete the hidden GNN nodes to maintain the architecture scalability.

\subsection{Summary}

Based on the above discussions, we briefly summarize the architecture-based methods in terms of fixed architecture and dynamic architecture. Their potential to reach continuous performance improvement is also discussed. 

\subsubsection{\textbf{Fixed Architecture}}

The fixed architecture method is limited by the model capacity and mainly focuses on parameter allocation and efficiency. It requires the model to identify important parameters and then release unimportant parameters for learning new tasks, which can be achieved through neural network search and uncertainty estimation, etc.

Due to the limitation of model capacity, the usable parameters will gradually decrease when more tasks are introduced. Therefore, parameter efficiency is another significant consideration. Current work usually imposes sparsity constraints on model parameters to ensure the parameter reusability and efficiency.

\subsubsection{\textbf{Dynamic Architecture}}

Dynamic architecture method can expand the model architecture dynamically. Compared with the fixed architecture method, it has a broader parameter space and a more flexible architecture learning strategy, like reinforcement learning and neural architecture search.

Although dynamic architecture has no concerns about model capacity, the high parameter efficiency is also important because low parameter efficiency will lead to the rapid expansion of model size, which decreases learning efficiency. There are several ways to ensure parameter efficiency besides the sparsity constraints, such as iterative pruning or freezing parameters.

\subsubsection{\textbf{Continuous Performance Improvement}}

Theoretically, there is also no knowledge enhancement in architecture-based methods, and it cannot reach continuous performance improvement. However, much work based on lottery tickets hypothesis \cite{frankle2019lottery} proves that there are some subnetworks (called winning tickets) that are much smaller than the entire network, which can achieve comparable or even better performance than entire network after training for enough iterations. An implicit but obvious fact is that the winning tickets vary with different subtasks. These winning tickets may share partial architectures and parameters. Besides, there may be multiple different winning tickets for a single subtask, and there exist unbiased winning tickets even in the networks that focus on spurious correlations \cite{zhang2021can}. These winning tickets of a single subtask are differential in generalization ability. Architecture-based methods are essential to finding an optimal public subnetwork, called lifelong ticket in \cite{chen2021long}, for all tasks. Lottery tickets theory also indirectly proves that different network structures also impact task performances. This gives us a theoretical possibility to find a set of lifelong tickets that can achieve continuous performance improvement. Besides, leveraging additional components, modules, or networks to transfer knowledge between different architectures may also probably achieve continuous performance improvement.

\section{Representation-based Methods}\label{sec:rep}

\begin{figure}[htbp]
\centering
\begin{forest}
    for tree={
      draw, semithick, rounded corners, drop shadow,
            align = left,   
             edge = {draw, semithick},
    parent anchor = east, 
     child anchor = west,
     		 grow = 0, reversed, 
            s sep = 1mm,    
            l sep = 3mm,    
            font  = \small
               }
[Representation, fill=red!30
    [Embedding Separation, fill=blue!30
	[Segmentation Basis, fill=green!30
		[Semanteme: \\ DiCGRL \cite{kou2020disentangle}, fill=orange!30]
		[Prototype: \\ HPN \cite{zhang2022hierarchical}, fill=orange!30]
	]
        [Updating Strategy, fill=green!30
            [Semantic-sharing Neighbors \\ Activation: DiCGRL \cite{kou2020disentangle}, fill=yellow!30]
            [Prototype Matching: \\ HPN \cite{zhang2022hierarchical}, fill=yellow!30]
        ]
    ]   	
    [Knowledge Transmission, fill=blue!30
        [Knowledge Type, fill=cyan!30
            [Task Correlations: \\ CML-BGNN \cite{luo2020learning}, fill=purple!30]
            [Temporal Information: \\ ICGN \cite{xia2021incremental}, fill=purple!30]
            [Long-term Preference: \\ CIGC \cite{ding2022causal}, fill=purple!30]
        ]
        [Transmission Medium, fill=cyan!30
            [RNN: \\ CML-BGNN \cite{luo2020learning}, fill=pink!30]
            [CNN: \\ ICGN \cite{xia2021incremental}; CIGC \cite{ding2022causal}, fill=pink!30]
        ]
    ]
]
\end{forest}
\caption{Representation-based CGL methods}
\label{representation}
\end{figure}
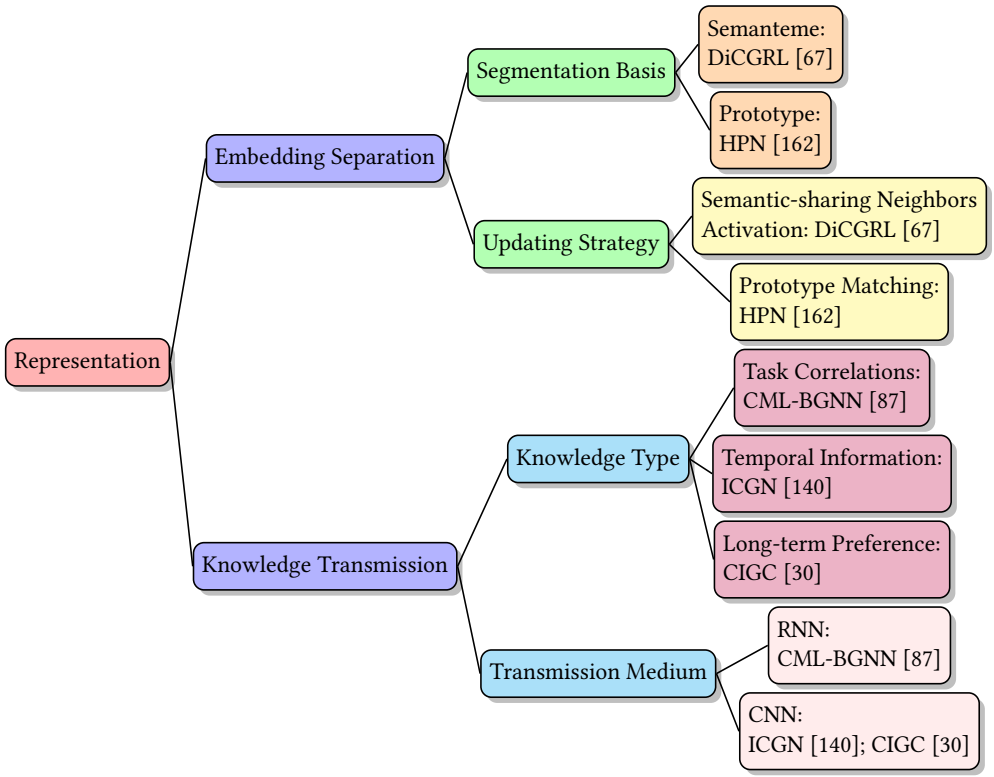

Due to the interdependence of nodes and the existence of inter-task edges, the new incremental graphs will impact previous graphs, and the knowledge of previous tasks can also be transferred to subsequent tasks. The knowledge is difficult to incorporate into continual graph learning explicitly but implicitly encoded in node embeddings, which we refer to as representation-based methods. The underlying principle is that the existing node embeddings already contain all the necessary information for downstream tasks, and the embeddings obtained by this method are equivalent to that of the weakened joint training. Generally, representation-based methods can be summarized into separation and transmission, as shown in Figure~\ref{representation}.

\subsection{Embedding Separation}

Embedding separation methods typically identify the influenced components and uninfluenced ones of graph embeddings first and then separate the embeddings for differentiated processing to update representations. There are two issues for separation methods: separation basis (minimum structural component of separation), and updating strategy (how the embeddings update).

\subsubsection{\textbf{Segmentation Basis}}

There are several ways to separate embeddings into several components. The semantic-based methods achieve component separation through semantic segmentation of graph embeddings. DiCGRL \cite{kou2020disentangle} decouples the relational triplets in the graph into multiple independent components according to their semantic aspects, and leverages graph embedding methods to learn separated graph embeddings.

Another way is based on prototypes. HPN~\cite{zhang2022hierarchical} proposes to extract different levels of abstract knowledge in the form of prototypes. Specifically, each node embedding is represented by three levels of prototypes.

\subsubsection{\textbf{Updating Strategy}}

The updating strategy specifies how the related components of embeddings are updated. For semantic-based separation, DiCGRL \cite{kou2020disentangle} adopts a semantic-sharing neighbors activation strategy. Specifically, the relational triplets from previous tasks that need to be updated are first identified and their neighbors are also affected. To reduce computation costs, each triplet only activates the neighbors with related semantic components. At last, only the common semantic components are updated through graph embedding methods.

In HPN \cite{zhang2022hierarchical}, three types of prototypes are used to represent the graph embeddings. The increasing graph data is extracted into atomic embeddings, which are either matched to existing atomic prototypes or used to build new atomic prototypes. The node embeddings and class embeddings are then matched to existing node prototypes and class prototypes or used to establish new node prototypes and class prototypes. Therefore, the new prototypes will be established when the new knowledge cannot match the existing prototypes.

\subsection{Knowledge Transmission}

Knowledge transmission methods transmit the previous experience and knowledge to the model when learning new tasks. The key to the methods is the type of knowledge and the way of transmission. 

\subsubsection{\textbf{Knowledge Type}}

Besides the knowledge from the task dataset itself, other knowledge may be also important for task learning. For example, task correlations \cite{luo2020learning}, temporal information \cite{xia2021incremental}, and long-term preference \cite{ding2022causal}. CML-BGNN \cite{luo2020learning} formulates meta-learning as a continual graph learning task, in which each task is modeled as a graph and each sample from the query sets and the support sets is modeled as a node. It preserves the intra-task interactions of the nodes and also the inter-task correlations to transfer them to the next tasks.

IGCN \cite{xia2021incremental} believes temporal information is significant for collaborative filtering. It conserves the temporal information from the last period and then fuses it with that in the current period incrementally to achieve efficient and accurate temporal-aware feature learning.

CIGC \cite{ding2022causal} believes the long-term preference signal is important in recommender systems and preserves it through aggregating old representations with new ones.

\subsubsection{\textbf{Transmission Medium}}

Another issue is the transmission way. Current work mainly transfers knowledge through RNN and CNN. CML-BGNN \cite{luo2020learning} uses gated recurrent units to transfer the history information and task information to the next tasks. The long-term task correlations are captured and transferred.

Instead, ICGN \cite{xia2021incremental} uses CNN to transfer knowledge because the RNN-based method suffers from catastrophic forgetting problems when updating user-item embeddings. CIGC \cite{ding2022causal} also adopts a simplified CNN design to aggregate and fuse old and new representations to overcome catastrophic forgetting.

\subsection{Summary}

We simply summarize the representation-based methods in this part.

\subsubsection{\textbf{Embedding Seperation}}

Embedding Separation methods follow the idea that graph embeddings are composed of several individual components. These components are combined in a parallel \cite{ding2022causal} or a hierarchical \cite{zhang2022hierarchical} way and can be selectively and differently updated. Besides the semantic-based and prototype-based separations, other ways to disentangle the graph embeddings remain to be explored. 

Selectively updating the components requires an appropriate updating strategy. Typically, parallel combinations identify the embeddings that need to be updated and then update them as well as their neighbors who share common components with them. Hierarchical combinations update different levels of components when the current combinations cannot match the new knowledge.

\subsubsection{\textbf{Knowledge Transmission}}

Knowledge Transmission methods mainly use additional networks or modules to transfer the experience and knowledge from the previous tasks to the next tasks. These experiences or knowledge such as task correlations, temporal information, or long-term signals are usually implicit. It requires the identification of important knowledge, which depends on the specific application scenarios and tasks.

Transmission mediums are mainly RNNs and CNNs. RNNs such as GURs and LSTM can capture sequential related knowledge but suffer more from forgetting and they need new inputs at every time step to transfer knowledge. CNNs and related variants specialize in capturing spatial relationships. The selection of transmission medium also depends on the specific tasks and scenarios.

\subsubsection{\textbf{Continuous Performance Improvement}}

This family of methods inherently implies knowledge enhancement through representation updating, and therefore is theoretically guaranteed for continuous performance improvement. A typical example is the prototype theory-based methods \cite{qin2020perpetual,zhang2022hierarchical}, in which the learning is based on prototypes and the learning of new tasks is the combination representation of the existing prototypes.

\section{Applications}\label{sec:app}

Graph learning has been widely applied in various areas and there are some widely used datasets. \cite{hu2020open} provides a benchmark that contains several graph datasets for graph learning. The datasets for continual graph learning are built from the general datasets used for graph learning. Some recent efforts \cite{zhang2022cglb,ko2022begin} have also been made to build such datasets. Nevertheless, these open datasets built for continual graph learning still cannot cover the scope involved in the current continual graph learning studies. In this part, we first summarize the datasets used in current continual graph learning studies and their corresponding applications, as Table~\ref{application} presents. Then we summarize the areas to which these datasets belong and discuss some possible applications of continual graph learning in these areas.

\begin{table*}[t]
\caption{Datasets and Applications}
\begin{center}
\begin{tabular}{p{2cm}p{5.5cm}p{3cm}m{2.2cm}}

\toprule
 Areas & Datasets & Applications & References \\
 
\midrule

\makecell[l]{Natural \\ Language \\ Processing} & \makecell[l]{FB15K-237, WN18RR} & \makecell[l]{Knowledge graph \\ completion} & \cite{kou2020disentangle} \\
\hline

\makecell[l]{Computer \\Visions} & \makecell[l]{CIFAR-100, ImageNet, miniImageNet,  \\tieredImageNet, CUB200} & \makecell[l]{Image classification} & \cite{dong2021few,luo2020learning,liu2021structural} \\
\hline

\makecell[l]{Recommender \\Systems} & \makecell[l]{Yelp, Gowalla, Adressa, Netflix,\\ Taobao, LastFM, Amazon Books, \\ MovieLens, Reddit, Actor, Flickr, \\ OGB-Products, Amazon-Products, \\ Coauthors-CS, Amazon-Computers}  & \makecell[l]{Top-K \\recommendation} & \cite{ding2022causal,xia2021incremental,ahrabian2021structure,wang2021graph,xu2020graphsail,zhang2022hierarchical,kim2022dygrain,perini2022learning,zhou2021overcoming,rakaraddi2022reinforced,liu2021overcoming,su2023towards,cai2022multimodal,lu2022geometer,wang2021graph,sun2023self} \\
\hline

\makecell[l]{Citation \\Networks} & \makecell[l]{Cora, CiteSeer, PubMed, DBLP, \\ OGB-Arxiv} & \makecell[l]{Paper classification} & \cite{kou2020disentangle,zhang2022hierarchical,kim2022dygrain,perini2022learning,zhou2021overcoming,rakaraddi2022reinforced,wang2020streaming,wang2022streaming,su2023towards,liu2021overcoming,lu2022geometer,sun2023self} \\
\hline

\makecell[l]{Bioinformatics} & \makecell[l]{Tox21, PPI} & \makecell[l]{Protein property/\\ structure prediction} & \cite{liu2021overcoming} \\
\hline

\makecell[l]{Transportation \\ Systems} & \makecell[l]{PEMSD3-Stream} &  \makecell[l]{Traffic flow \\ forecasting} &  \cite{chen2021trafficstream} \\
\hline

\makecell[l]{Financial \\ Systems} & \makecell[l]{Elliptic} & \makecell[l]{Fraud detection} & \cite{wang2020streaming,wang2022streaming,perini2022learning} \\

\bottomrule

\end{tabular}
\label{application}
\end{center}
\end{table*}

\subsection{Natural Language Processing}

Although text inputs are usually expressed as a sequence of tokens, the structural information between texts like syntactic parsing trees can be exploited to enhance original texts. Texts can be modeled as various graphs, such as dependency graphs, constituency graphs, abstract meaning representation graphs, information extraction graphs, discourse graphs, knowledge graphs, etc. \cite{wu2023graph}, which gives deep graph models wide applicability in NLP tasks. This graph-based representation allows for capturing semantic and structural relationships between words, sentences, or documents, achieving better performance in NLP applications, including natural language generation, machine reading comprehension, question answering, dialog systems, etc., as summarized in \cite{wu2023graph}.

Due to the openness, diversity, flexibility, contextual dependencies, and evolution of natural languages, continual learning is imperative for NLP. For example, a large language model needs continuous updating. A QA system also needs to learn continuously to adapt to the dialogue with users \cite{biesialska-etal-2020-continual}. Current continual graph learning in NLP mainly focuses on knowledge graphs, especially on knowledge graph completion, among which GNNs are mainly used as graph encoders.

\subsection{Computer Visions}

Although computer vision tasks are generally handled by CNNs, some recent works have also shown the feasibility of GNNs in computer vision tasks, such as ML-GCN \cite{chen2019multi}, Curve-GCN \cite{ling2019fast}, and Vision GNN \cite{han2022vision}. GNNs have been applied in various fields of CV, such as 2D natural images, videos, multi-modality of vision and language, 3D data, and medical images. More specific tasks of GNNs in CV can be referred to \cite{chen2022survey}. The key to GNNs for performing computer vision tasks is how the graph is built from regular grid-like structures. Some recent work constructs graphs by modeling the topological structure between features \cite{lei2020class}, exemplars \cite{dong2021few}, and tasks \cite{luo2020learning}. 

Currently, most continual learning work is conducted within CV tasks. Nevertheless, the applications of continual graph learning in CV are still in their infancy. Existing continual graph learning work in CV usually constructs graphs from images to preserve structural knowledge for continual learning, such as exemplar relation graph \cite{dong2021few} and memory knowledge graph \cite{liu2021structural}. It is believed that continual graph learning in this area will be more focused on in the future.

\subsection{Recommender Systems}

There are many graph structures in recommender systems, such as user-item bipartite graphs, user behavior sequence graphs, user social relationship graphs, product co-purchasing graphs, and item knowledge graphs. GNNs are widely used in recommender systems as they can improve the representations of users and items by explicitly encoding topological structures with multi-hop relationships through node aggregations, with additional information from social relationships or knowledge graphs. These improved user representations are then used for various applications in recommender systems, such as user-item collaborative filtering, sequential recommendations, social recommendations, and knowledge graph-based recommendations. Besides, GNNs are also applied in other recommender tasks such as click-through rate (CTR) prediction, point-of-interest (POI) recommendation, group recommendation, and bundle recommendation. More applications of GNNs in recommender systems can be referred to \cite{wu2022graph,gao2023survey,sharma2022survey}.

Continual learning is essential to recommender systems, especially e-commerce recommender systems, because they are naturally dynamic scenarios, in which users' preferences and items' features may change over time. Current continual graph learning work on recommender systems mainly focuses on top-$k$ recommendations \cite{wang2021graph,xu2020graphsail,ahrabian2021structure,ding2022causal,xia2021incremental}.

\subsection{Citation Networks}

A citation network can be constructed by treating papers as nodes and their citation relationships as edges. It evolves over time and continual learning is essential for citation network-related tasks. Since citation networks are highly homophilic, they are often adopted as benchmark datasets by many GNNs for node classification. Many current continual graph learning works use some public citation networks for node classification tasks. Besides, GNNs can be also applied to other applications in citation networks, including citation prediction, co-author prediction, influence analysis, and visualization. 

Paper classification classifies papers into different subject areas, topics, or types. Citation prediction predicts the citing behavior of papers, for example, whether a paper will be cited or which other papers a given paper will cite. The co-author prediction aims to predict the co-authorships that exist between specific scholars in which fields or topics. The academic influence of specific scholars, institutions, or fields can be analyzed by GNNs. For instance, predicting the academic achievements of certain scholars or the development trends of specific fields. Also, the node and edge embedding vectors can be obtained for visualizing citation networks, which helps people better understand knowledge dissemination and impact distribution issues within academia.

\subsection{Bioinformatics}

In bioinformatics, molecules, and chemical compounds can naturally be represented as graphs with atoms as nodes and chemical bonds as edges, which enables the applications of GNNs in this area, including drug discovery, property/structure prediction, etc. In drug discovery, drugs with similar pharmacological activities usually have similar molecular structures. GNN-based methods like molecular representation learning, and molecular graph generation can predict chemical properties and activities on a large scale effectively by considering the molecular structures internally, speeding up the drug discovery process. GNN-based models have also been applied in medical diagnosis \cite{ahmedt2021graph,liu2020identification,huang2020edge}. 

Continual learning in biochemistry and biomedicine is still rarely investigated. A model with continual learning ability is meaningful for bioinformatics. For example, a drug discovery model may continually discover targeted molecules. A disease diagnosis model may identify new diseases or viruses with continual learning. Continual graph learning provides a promising direction.

\subsection{Transportation Systems}

A transportation network can be modeled as a graph by considering the locations or points of interest (intersections, bus stops, parking lots) as nodes, and the links between these nodes as edges. GNNs can be applied in many transportation applications like traffic forecasting, demand prediction, autonomous vehicles, intersection management, parking management, urban planning, and transportation safety. More information can be referred to \cite{rahmani2023graph,ye2020build}. Current work mainly focuses on traffic forecasting.

A transportation network is considered to be dependent both in the spatial and temporal domains since the flow and speed of a road depend on the flow and speed of its neighbors. Therefore, GNNs, especially spatial-temporal GNNs \cite{yu2018spatio}, are widely used for traffic network analysis. Most studies assume that the dependency of two road segments is considered static and represented by a predefined or self-learning adjacency matrix. Continual graph learning is applicable for transportation systems since they evolve over time, enabling GNNs to cope with the evolution of transportation networks without forgetting previous patterns.

\subsection{Financial Systems}

Graphs can be constructed in financial systems, like transaction networks, user-item review graphs, and stock relation graphs. GNNs are widely used in this domain, including stock movement prediction, loan default risk prediction, fraud detection, and event prediction. \cite{wang2022review} summarizes the applications of GNNs in the financial domain.

Apparently, financial systems are also dynamic, and continual learning is important for financial investment and security, either in traditional financial systems or in cryptocurrencies. The financial transactions are filled with illegal activities such as fraud and money laundering. Identifying illicit activities like fraud and money laundering in financial transaction networks is significant, especially in cryptocurrencies like Bitcoin and Ethereum. There have been many studies on fraud detection \cite{singh2021temporal,jin2022heterogeneous} and anti-money laundering \cite{lo2023inspection} in cryptocurrencies. Continual graph learning is helpful for detecting illicit addresses and transactions in evolving transaction networks.

\section{Open Issues}\label{sec:open}

Essentially, most current continual graph learning methods are compromises and trade-offs between plasticity and stability. Although catastrophic forgetting has been effectively addressed, continuous performance improvement has not been systematically investigated, despite isolated efforts (e.g., HPN \cite{zhang2022hierarchical}) achieving positive backward transfer. Additionally, positive backward transfer only means $BWT=\frac{1}{T-1}\sum_{j=1}^{T-1}R_{T,j}-R_{j,j}>0$, while continuous performance improvement requires that $R_{T,j}>R_{T-1,j}$ be true all the time, which is more demanding than positive backward transfer.

Focusing on continuous performance improvement, we outline some considerations that may impact the achievement of this objective and raise some open issues and corresponding future directions.

\subsection{Convergence of Continual Graph Learning}

When will the models converge is an unavoidable problem when discussing continuous performance improvement. While knowledge enhancement and optimization controlling provide a roadmap for achieving continuous performance improvement, realizing this objective may necessitate the imposition of more stringent conditions, for example, sufficient model capacity.

Theoretically, a neural network with fixed architecture is limited in capacity. Continual graph learning essentially finds a shared parameter solution for all tasks, and the complexity of finding a desirable solution mainly depends on the structure of parameter space. Learning new tasks in continual graph learning settings is equivalent to learning new tasks in a limited and constantly shrinking parameter space while overcoming catastrophic forgetting. However, it is proved that this problem is generally NP-hard \cite{knoblauch2020optimal} since the feasible parameter space tends to be irregular and narrow as more tasks are introduced. Eventually, the model will inevitably fail to learn new tasks under such conditions.

Practically, the number of learning tasks can be extended by expanding the network architecture, which allocates task-specific parameters without sharing parameters between tasks and gets rid of the limitation of model capacity. This practice also possibly achieves continuous performance improvement. Nevertheless, it may restricted by the upper bound of continual graph learning performance determined by the joint training for joint optimization of all tasks. Ideally, the model converges when the performance reaches or approximates this upper bound. However, the question is how long will it take for the performance to reach this upper bound, and how long can it last after reaching it?

Future directions on convergence may include two aspects. One is how the performance can reach the upper bound as soon as possible and how the model maintains the best performance for as long as possible. The other is how the number of tasks learned by a model can be expanded as much as possible.

\subsection{Scalability}

Continuous performance improvement is also affected by the number of tasks. A continual graph learning model is expected to learn as many tasks as possible. However, as discussed previously, as more tasks are introduced, the feasible parameter space tends to shrink quickly and it is an NP-hard problem to find a shared solution for all tasks. Therefore, the challenge of achieving continuous performance improvement will escalate with the increasing number of tasks. An intuitive idea is to increase the data size of each task as a means of reducing the number of tasks. 

However, the graphs in the real world are usually large-scale and highly complex. For example, a social network may have tens of millions or even billions of nodes and edges, making scalability a great challenge for CGL, despite the continual graph learning settings splitting the graph into multiple tasks and reducing the memory requirement for each training. However, as discussed previously, as more tasks are introduced, the feasible parameter space tends to shrink quickly and it is an NP-hard problem to find a shared solution for all tasks. Therefore, there is a dilemma between the task number and the data size of each task. 

A typical way to reduce the size of a graph is sampling, like the neighbor sampling in GraphSAGE \cite{hamilton2017inductive} and random walk sampling in PinSage \cite{ying2018graph}. However, sampling will inevitably lead to information loss, which may result in performance degradation. Another option is to design and develop graph algorithms that can handle large graphs effectively, like simplified GCNs \cite{he2020lightgcn,wu2019simplifying}, which are more scalable due to the simplifications. Nevertheless, these simplified models are limited by the choices of specific components like aggregators or updates.

\subsection{Robustness}

The robustness of a CGL model also impacts continuous performance improvement. The robustness of a model refers to its ability to resist noise, outliers, or attacks in input data. There are usually some noise and outliers in graphs in real-world applications. The attackers may also intentionally add noise or attack the models to disrupt their performance. Moreover, the message-passing pattern in GNNs will make the prediction of nodes heavily dependent on specific neighboring nodes, making GNNs poorly tolerant of noise. For example, \cite{zugner2018adversarial} shows that attacking the target node indirectly by changing the attributes of its neighbors can achieve the goal of attacking the target node. In CGL, the strong robustness of GNNs is equally indispensable, which can avoid noise disturbances to model performance. Therefore, corresponding efforts should be invested in research on strong robustness, such as graph data enhancement, adversarial attacks, and defenses on graphs.

\subsection{Privacy Preservation}

Data availability is another consideration for continuous performance improvement. Many real-world data like social networks, medical data, and financial networks have strict requirements for privacy preservation. In most cases, these data cannot be collected for continual graph learning. Therefore, how continual graph learning can be conducted under the premise of preserving privacy becomes a great challenge. A typical paradigm is graph federated learning where the data does not need to be uploaded into a center server. However, isolated graphs in horizontal intra-graph federated learning is a huge challenge. Local subgraphs only contain lower-order neighbor information and cannot capture the higher-order neighbor information from other subgraphs. Besides, the communication consumption between the server and clients also hinders the efficiency of graph federated learning.

Another way to ensure privacy preservation is differential privacy with graph learning, in which arbitrary small changes such as additions or deletions of nodes and edges and their features do not change the statistics, and thus any information cannot be inferred from the statistics. Graph learning can be performed with center differential privacy or local differential privacy. However, differential privacy usually decreases performance. There is a growing demand for privacy preservation in today's society. Continual graph learning with privacy preservation would be a promising direction.

\subsection{Unsupervised Continual Graph Learning}

Label sparsity likewise exerts an influence on continuous performance improvement. In the real world, most of the nodes on graphs are unlabeled and most current studies on GNNs are semi-supervised. However, semi-supervised graph learning relies heavily on labels and results in poor generalization and weak robustness \cite{zheng2022rethinking}. A promising way is unsupervised graph learning, especially self-supervised learning. Self-supervised graph learning can learn better graph representations by augmenting graphs through various methods like contrastive learning \cite{zeng2021contrastive} or self-generated labels \cite{wu2023self}.

In continual graph learning settings, self-supervised learning can leverage additional knowledge like mutual information, which may enhance performance. Besides, better graph representations may also improve the generalization ability of GNNs and therefore reduce the difficulty of finding a shared solution in a parameter space.

\subsection{Explainability}

Explainability is of significance to continual performance improvement. On the one hand, good explainability helps the graph models understand what knowledge is important for learning tasks, which helps graph models retain the important knowledge. On the other hand, satisfactory explainability also helps graph models to know why the learned knowledge can overcome catastrophic forgetting and even achieve continuous performance improvement.

There have been some efforts of explainability on GNNs \cite{miao2022interpretable,huang2022graphlime} and continual learning \cite{rymarczyk2023icicle}. The explanations on GNNs \cite{yuan2022explainability} can be categorized into instance-level and model-level methods, where the instance-level methods provide input-dependent explanations for each input graph by identifying important features for its prediction, model-level methods provide high-level insights and a generic understanding. However, the explainability of continual graph learning remains to be investigated.

\subsection{Continual Learning for Large Graph Models}

Artificial general intelligence (AGI) represented by large language models (LLMs) has achieved amazing performance in language-related tasks. LLMs-driven AI agents are expected to learn and perform like human beings. However, current LLMs have neither the ability of continual learning nor the large graph models (LGMs), although LGM is still in its infancy and there are no widely recognized LGMs yet. 

Nevertheless, an LGM is expected to have the desired ability to handle all different graph tasks across various domains and continuously improve the performance as the model size, dataset size, and training computation increase, which is summarized as graph scaling laws, graph foundation models, in-context graph understanding and processing, and versatile graph reasoning capabilities~\cite{zhang2023large}. Although recent efforts have been devoted to graph transformers \cite{ying2021transformers,rampavsek2022recipe}, graph pre-training paradigm \cite{liu2022graph,wu2023self}, and LLMs as graph models \cite{wang2023can,guo2023gpt4graph} to explore the potential of LGMs, the ability of continual learning emerges as a pivotal aspect to imbue LGMs with the aforementioned desirable characteristics.

\section{Conclusion}\label{sec:conc}

Owing to the ubiquity and dynamic nature of graphs in the real world, graph models represented by GNNs have been widely used in various fields. Continual graph learning is a newly emerging learning paradigm that aims to conduct graph learning tasks in continual learning settings and achieve continuous performance improvement. In this survey, we provide a comprehensive review of the recent studies on continual graph learning. A new taxonomy of continual graph learning methods that overcome catastrophic forgetting is proposed. Moreover, for each category, we briefly clarify the key issues, detail the corresponding practices in current studies, and discuss the possible solutions to achieve continuous performance improvement. Furthermore, we also propose some open issues related to continual performance improvement and suggest corresponding promising research directions. We hope this survey can help readers understand the recent progress in continual graph learning and shed light on the future development of this promising field.

\begin{acks}
This work was supported by the Macao Science and Technology Development Fund through the Macao Funding Scheme for Key Research and Development Projects (Grant Number: 0025/2019/AKP).
\end{acks}

\bibliographystyle{ACM-Reference-Format}
\bibliography{reference}

\end{document}